\documentclass{article}
\usepackage{iclr2027_conference,times}

\usepackage[T1]{fontenc}
\usepackage{microtype}
\usepackage{amsmath,amssymb}
\usepackage{booktabs}
\usepackage{array}
\usepackage{graphicx}
\usepackage{pgfplots}\pgfplotsset{compat=1.16}
\usepackage{xcolor}
\usepackage{float}
\usepackage[hidelinks]{hyperref}

\usepackage{caption}
\newcommand{\dfa}{\textsc{dfa}}
\newcommand{\modk}{\texttt{mod\_k}}

\providecommand{\Description}[2][]{}
\newcommand{\appendixnote}{}
\newcommand{\twocol}[2]{#2}
\newenvironment{widefigure}[1][t]{\begin{figure}[#1]\centering}{\end{figure}}
\newenvironment{widetable}[1][t]{\begin{table}[#1]\centering\small}{\end{table}}
\newenvironment{catalogtable}{\begin{table}[H]\centering\small}{\end{table}}

\title{\normalfont\raggedright When Does Reward Teach State?
A Hidden-Automaton Instrument and a Group-Language Warning Signal}

\author{James E. Allchin\\Independent Researcher}

\iclrfinalcopy % de-anonymize (show authors); the venue header is blanked below for the preprint.

\begin{document}
\maketitle
% \maketitle sets the style's venue running header; clear it before page 1
% ships (fancyhdr reads the header at shipout) so the preprint claims no venue.
\lhead{}
\begin{abstract}
Does a reinforcement-learning agent that earns reward learn its task's hidden state?
We study this question with hidden finite automata that the agent partially controls. Because each
automaton is known, we can normalize reward by the best achievable return and probe the network for
the true state at every step. Together the two measurements separate failures that reward alone
conflates. An agent can encode too little of a state its network could hold, or encode the state
and still control poorly. Weak on-policy RL matches random play while the state
probe stays at chance. State learning depends on the optimizer, the training budget, and the task's
structure. Permutation automata provide a warning before training: no input symbol maps two
distinct states to the same successor. On a stratified held-out set, 86 of 103 permutation automata
fail the state probe, and the classification is stable across probe read-outs and recovery
thresholds. Most of these failures come with weak reward. High reward without the state occurs but
is rare. Non-permutation automata can also fail. Oracle-normalized reward alone therefore does not
establish that the task's state was learned.

\end{abstract}

\section{Introduction}
Many agent tasks are finite-state. Examples include onboarding and support workflows, tool-use and
API-call sequences, compliance and triage procedures, and multi-step form repair. An agent can look
competent in training and then fail when the workflow is longer, noisier, or drawn from a different
distribution. It may have learned a reward-seeking shortcut instead of the underlying state machine
\citep{geirhos2020shortcut}, a form of goal misgeneralization \citep{langosco2022,shah2022}. Earlier
demonstrations have used behavior and an informal definition of the ``true goal.'' Here we know the
automaton's state at every step. We can normalize reward and test whether the network encodes that
state.

We turn formal-language recognition into a control problem: can an agent steer a hidden automaton
toward an accepting state, and does it come to represent that automaton's latent state while doing
so? Reward success and latent-state learning are then two separately measurable axes.

\textbf{What is hidden.} The agent sees every emitted symbol. Only the automaton's state is
hidden, and that state is a deterministic function of the observed history. There is no
irreducible uncertainty to integrate, so this is not POMDP belief-tracking. The question is whether
the agent can compute and remember a known function of the observed history.

\textbf{Thesis.} Reward success and latent-state learning are separate measurements. Their
relationship depends on three factors (Figure~\ref{fig:axes}): optimizer strength, whether the
automaton is a \emph{permutation automaton} (no symbol merges states), and how much the observations
reveal about the latent state.
Weak RL leaves both at failure: random-play return and a chance probe.
A stronger optimizer restores the coupling only partially.
A label-free prediction loss recovers state only as observations become informative.

\begin{widefigure}[t]
\includegraphics[width=\linewidth]{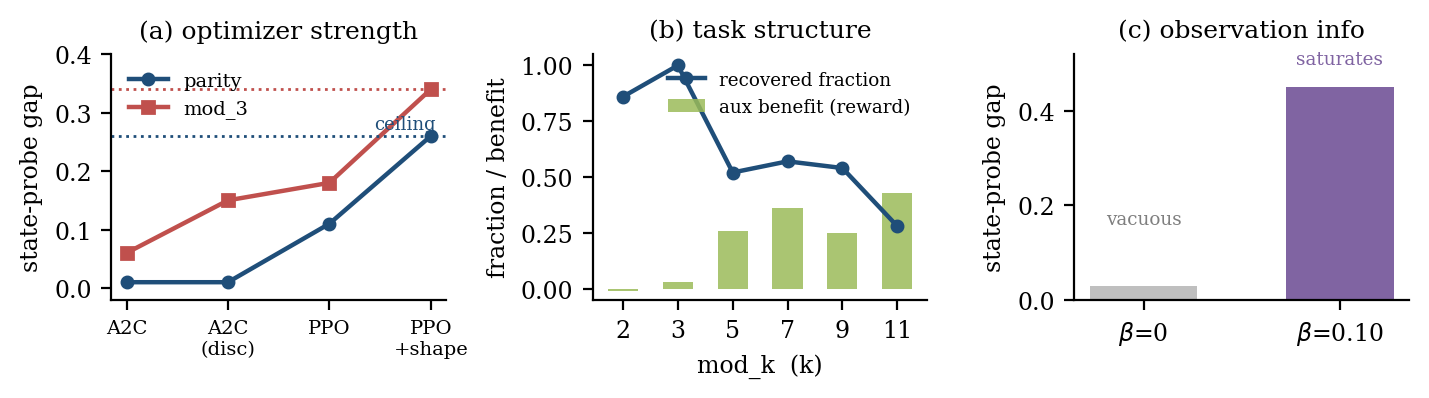}
\Description{Three side-by-side panels: probe gap versus optimizer competence, state recovery and
auxiliary benefit versus counter modulus, and probe gap versus observation informativeness.}%
\caption{\textbf{Three axes of the reward$\leftrightarrow$state coupling}, replotted from
\S\ref{sec:opt}--\S\ref{sec:scale}. \textbf{(a)} Probe gap (held-out state-probe accuracy over
the majority-class baseline) vs.\ optimizer competence, closing once reward shaping from the
oracle's value supplies credit directly. \textbf{(b)} As the \modk{} counter
grows, sparse PPO recovers a shrinking share of the recoverable state. \textbf{(c)} A
label-free prediction auxiliary adds nothing at $\beta{=}0$ and saturates by $\beta{=}0.10$.}
\label{fig:axes}
\end{widefigure}

\medskip
\textbf{Contributions.}
{\setlength{\topsep}{2pt}\setlength{\partopsep}{0pt}%
\begin{itemize}\setlength{\itemsep}{2pt}\setlength{\parskip}{0pt}
\item A measurement method for the question ``did the agent learn the task state, or a reward
shortcut?'' The known automaton supplies three measurements: reward scored against the
optimal return, a linear probe of the encoded state, and a supervised off-policy bound on
what the architecture could represent. We release the environment (\texttt{PartialControlDFAEnv}) with a taxonomy
separating diagnostic automata from trivial ones.
\item An optimizer-strength account of the reward$\leftrightarrow$\twocol{\allowbreak}{}state coupling. Under
weak on-policy RL, return stays at the random-play baseline and the state probe stays at chance.
A stronger optimizer couples the two only partially. A pre-specified A2C-vs-PPO control shows the
optimizer matters more than the credit scheme. We also analyze a privileged auxiliary that predicts
the true state and deployable label-free auxiliaries that predict upcoming observations.
\item A pre-training warning signal for a \emph{perception gap}: the network could represent the
state, but the trained policy does not reach the state-recovery threshold
(\S\ref{sec:predictor}).
When the automaton is a permutation (group-language) automaton, a property readable from the
transition function before any training, the agent rarely learns the state in our experiments. A
credit-assignment analysis gives a candidate mechanism (Appendix~\ref{app:credit}).\appendixnote{} Held out on
$153$ capacity-controlled fresh automata, $86$ of $103$ permutation automata fail the state probe. Informative observations with label-free prediction close the gap, and a hold action mitigates it (Appendix~\ref{app:validation}).
\end{itemize}}
The ground-truth probe separates a perception gap from a \emph{planning gap}, where the
state is decodable and reward stays low. Reward and the oracle alone cannot tell them apart.
A port of the instrument to a Qwen2.5-3B/7B LLM agent \citep{qwen2025}, run without chain of thought, is a
feasibility check in Appendix~\ref{app:llm}. The structure axis, the optimizer ordering, and
the lock-and-escape dynamics of \S\ref{sec:opt} appear there. A reward-successful perception gap does not. That port
is not part of the main claims.

\section{Related Work}\label{sec:related}
\textbf{Reward machines and automaton-structured RL} expose the automaton behind a (possibly
non-Markov) reward and let RL exploit it \citep{icarte2018}. A parallel line infers the
latent automaton, either to plan with \citep{icarte2022} or to model history-dependent dynamics as a
regular decision process via a finite transducer, with PAC guarantees \citep{brafman2019rdp}.
We hide the automaton and ask whether sparse-reward RL discovers its latent state as a byproduct
of maximizing reward. The known automaton is a measurement instrument, not a planner's input.
We study the opposite risk: the agent locks onto a reward-correlated shortcut, the control analog
of causal confusion in imitation \citep{dehaan2019}.

\textbf{Transformers, RNNs and counting.} Counting tasks such as parity separate architectures.
Recurrent nets count inductively. Transformers need positional tricks and tend not to
length-generalize \citep{hahn2020limits,strobl2024survey,chang2024count,huang2025lengthgen,deletang2023,bhattamishra2020,weiss2018}.
\citet{michalenko2019} decode RNN hidden states against minimal-DFA states, the recurrent analog
of our probe, for supervised recognition rather than sparse-reward control.
\citet{liu2022shortcuts} show low-depth transformers can simulate any finite automaton, with
brittle constant-depth shortcuts that degrade as supervision gets sparser. An LSTM stays at
$100\%$ on the same group automata. Our testbed puts that separation inside an RL control
loop with a hidden, sparsely-rewarded state, and gives it a normalized, probeable form.
The group automata in that comparison require a running count rather than a local pattern.
A last-bit machine is local; a cyclic counter with three or more residues is not.

\textbf{Emergent world representations and probing.} A model trained only to predict the next token
can represent the latent state of the generating process. Othello-GPT encodes the board, linearly
decodable and causally tied to predictions \citep{li2023othello,nanda2023linear}, with parallels in
chess \citep{toshniwal2022chess,karvonen2024chess} and program semantics \citep{jin2024programs}. Those models are trained by supervised next-token prediction on
expert play. We study sparse-reward RL under partial control, with an exact oracle and a known
finite probe target, and we drive reward and state apart. We adopt the probing safeguards of
this literature (control tasks and selectivity \citep{hewitt2019,alain2016}) and run an
interventional probe.

\textbf{State abstraction and auxiliary tasks.} Bisimulation and predictive
state representations define state through observable consequences rather than a
posited hidden variable \citep{littman2001psr,gelada2019,zhang2021dbc}.
Our \dfa{} state is known exactly and used for measurement, not learned as the agent's state.
Recurrent model-free RL is a strong POMDP baseline \citep{ni2022}. Auxiliary self-supervised
objectives are standard under sparse reward \citep{jaderberg2017unreal}. A loss that predicts the
true \dfa{} state has a known target, so an independent probe can check that the intended
representation was installed.

\textbf{Reward hacking.} Benchmarks measure exploit rates behaviorally
\citep{pan2022mis,skalse2022}. We add a white-box substrate where ``did it learn the task state''
is a probe reading, complementary to those rates.

\section{The Hidden-Automaton Control Testbed}\label{sec:testbed}

\begin{widefigure}[t]
\begin{tikzpicture}[font=\small,
  sym/.style={draw, minimum size=5.5mm, inner sep=1pt},
  env/.style={sym, fill=gray!12, draw=gray!60},
  agt/.style={sym, fill=blue!10, draw=blue!55!black, thick},
  st/.style={draw=none, inner sep=1.5pt}]
  \node[anchor=east] at (0.55,1.2) {observed symbols};
  \node[env] (a1) at (1.4,1.2) {1};
  \node[agt] (a2) at (2.8,1.2) {0};
  \node[env] (a3) at (4.2,1.2) {1};
  \node[env] (a4) at (5.6,1.2) {1};
  \node[env] (a5) at (7.0,1.2) {0};
  \node[agt] (a6) at (8.4,1.2) {1};
  \node[text=gray!50!black, font=\scriptsize, anchor=north] at (1.4,0.92) {env draws};
  \node[text=blue!55!black, font=\scriptsize, anchor=north] at (2.8,0.92) {agent picks};
  \node[text=blue!55!black, font=\scriptsize, anchor=north] at (8.4,0.92) {agent picks};
  \node[anchor=east] at (0.55,0) {hidden state};
  \node[st] (q0) at (0.7,0) {E};
  \node[st] (q1) at (2.1,0) {O};
  \node[st] (q2) at (3.5,0) {O};
  \node[st] (q3) at (4.9,0) {E};
  \node[st] (q4) at (6.3,0) {O};
  \node[st] (q5) at (7.7,0) {O};
  \node[st] (q6) at (9.1,0) {E};
  \foreach \i/\j in {q0/q1,q1/q2,q2/q3,q3/q4,q4/q5,q5/q6}{\draw[->,gray] (\i)--(\j);}
  \draw[dashed, rounded corners, gray!70] (0.38,-0.35) rectangle (9.42,0.35);
  \node[font=\scriptsize, text=gray!40!black, anchor=north west] at (0.38,-0.42)
    {never shown to the agent; the probe's read-out target};
  \node[anchor=west] at (9.5,0) {${\in}\,F \;\Rightarrow\; r{=}1$};
\end{tikzpicture}
\Description{A row of observed symbols, some drawn by the environment and some picked by the agent,
above the hidden-state sequence they drive; the hidden states are never shown to the agent.}%
\caption{One episode of the partial-control environment (\texttt{parity}, $T{=}6$). Each step emits
one symbol: the environment draws the gray ones at random and the agent picks the blue ones (a
$p_{\mathrm{control}}$ fraction of steps). Every symbol advances the hidden state, here the
parity of 1s (\textbf{E}ven/\textbf{O}dd). The agent sees only the top row; the only reward is terminal: $1$ if the
final state accepts.}
\label{fig:env}
\end{widefigure}

\textbf{Partial-control \dfa{} environment.} A task is a deterministic finite automaton: a set of
states $Q$, a symbol alphabet $\Sigma$, a transition function $\delta$ mapping a state and a symbol
to the next state, a start state $q_0$, and a set of accepting states $F$. An episode is a walk
through this automaton, $T$ symbols long.
At each step one symbol $a_t$ is emitted and the state
moves, $q_{t+1}=\delta(q_t,a_t)$. Who emits the symbol varies: with probability
$p_{\mathrm{control}}$ the agent chooses it, and otherwise the environment draws it uniformly at
random. The agent sees the emitted symbols, whether the current step is one it controls, and the
time remaining. It never sees the state $q_t$. Reward is sparse and terminal: $1$ if the
final state is accepting ($q_T\in F$), else $0$. For \texttt{parity}, for example, the state is the running
parity of the 1s emitted so far, and reward is $1$ only if their count ends even. The
agent's job is therefore to steer a partly random walk into an accepting state. Under full control
this is trivial for most tasks (emit an accepting word directly); under partial control the agent
must track the hidden state and spend its limited choices well (Figure~\ref{fig:env}).

\textbf{Exact oracle.} Because the automaton and the control probability are known, the best
possible play can be computed outright. Let $V(q,r)$ be the probability of ending accepted when
the current state is $q$ with $r$ steps remaining, the agent chooses optimally on the steps it
controls, and the environment plays randomly on the rest. With no steps left the outcome is
already decided, and one step of lookahead gives the backward induction:
\twocol{\begin{align*}
V(q,0)&=\mathbf{1}[q\in F],\\
V(q,r)&=p_{\mathrm{control}}\max_a V(\delta(q,a),r{-}1)\\
&\quad+(1{-}p_{\mathrm{control}})\,\mathbb{E}_a V(\delta(q,a),r{-}1).
\end{align*}}{\[
V(q,0)=\mathbf{1}[q\in F],\qquad
V(q,r)=p_{\mathrm{control}}\max_a V(\delta(q,a),r{-}1)+(1{-}p_{\mathrm{control}})\,\mathbb{E}_a V(\delta(q,a),r{-}1).
\]}
$V(q_0,T)$ is then the best achievable success rate for the whole task. We report
\emph{oracle-normalized success} (written val$_n$ below): the agent's empirical success rate
divided by $V(q_0,T)$, so $1.0$ is optimal. The same computation under always-random play gives
the random-play value of val$_n$ as $V_{\mathrm{random}}/V(q_0,T)$. On \texttt{parity} that
baseline is $0.67$; on \texttt{mod\_3} it is $0.48$. Always-agent play gives a full-control
reference. val$_n$ is then comparable across lengths and control probabilities within a task.
The diagnostic filter keeps random play well below optimal, so beating random is not an accident
of a trivial task.

\section{Diagnostic Instruments}\label{sec:instr}
\textbf{Linear \dfa-state probe (on-policy).} We train a linear classifier to predict the true
$q_t$ from the policy's features, using the agent's own rollouts, and report the held-out gap
over the majority-class baseline. We summarize a probe by its \emph{recovery}, the fraction of the
headroom above the majority class it attains,
$(\mathrm{acc}-\mathrm{maj})/(1-\mathrm{maj})$. Here $\mathrm{acc}$ is the probe's held-out
accuracy and $\mathrm{maj}$ is the accuracy of always predicting the most frequent state, so $0$
is chance and $1$ is a perfect read-out. Balanced accuracy is reported alongside to control class
imbalance, and a control-task/selectivity
check \citep{hewitt2019} constrains probe artifacts. An off-policy probe on uniform random play
separately tests visitation effects of the agent's own behavior
(Appendix~\ref{app:validation}).

\textbf{Interventional probe (causal).} We edit the activation along the decoded-state direction
and check whether the greedy action changes as $\delta$ predicts. That test measures how
sensitive the action is to the specified state-direction edit, read against a norm-matched
random direction.

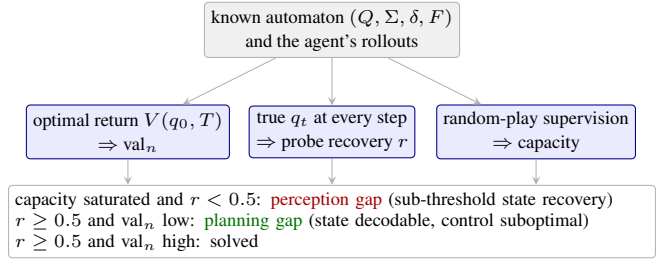
\begin{figure}[h!]\centering
\begin{tikzpicture}[font=\scriptsize, >=stealth,
  box/.style={draw, rounded corners=1.5pt, align=center, inner sep=2.5pt, minimum height=5.5mm},
  src/.style={box, fill=gray!12, draw=gray!60},
  meas/.style={box, fill=blue!8, draw=blue!55!black},
  verdict/.style={box, fill=white, draw=gray!60, text width=\twocol{0.9\linewidth}{0.6\linewidth}, align=left}]
  \node[src] (dfa) at (0,0) {known automaton $(Q,\Sigma,\delta,F)$\\[1pt]and the agent's rollouts};
  \node[meas] (rew) at (-2.7,-1.35) {optimal return $V(q_0,T)$\\[1pt]$\Rightarrow$ val$_n$};
  \node[meas] (pr) at (0,-1.35) {true $q_t$ at every step\\[1pt]$\Rightarrow$ probe recovery $r$};
  \node[meas] (cap) at (2.7,-1.35) {random-play supervision\\[1pt]$\Rightarrow$ capacity};
  \draw[->, gray!60] (dfa) -- (rew);
  \draw[->, gray!60] (dfa) -- (pr);
  \draw[->, gray!60] (dfa) -- (cap);
  \node[verdict, anchor=north] (v) at (0,-2.05) {%
    capacity saturated and $r<0.5$: \textcolor{red!65!black}{perception gap} (sub-threshold state recovery)\\
    $r\ge0.5$ and val$_n$ low: \textcolor{green!45!black}{planning gap} (state decodable, control suboptimal)\\
    $r\ge0.5$ and val$_n$ high: solved};
  \draw[->, gray!60] (rew) -- (rew |- v.north);
  \draw[->, gray!60] (pr) -- (v);
  \draw[->, gray!60] (cap) -- (cap |- v.north);
\end{tikzpicture}
\Description{Flow diagram: the known automaton and the agent's rollouts feed three read-outs
(optimal return giving normalized success, the true state giving probe recovery, random-play
supervision giving capacity), which combine into a verdict of perception gap, planning gap, or
solved.}%
\caption{The instrument: three read-outs and their verdict. Reward alone sees only the first.}
\label{fig:instrument}
\end{figure}

\textbf{Supervised capacity (off-policy).} We train each encoder with full supervision to predict
$q_t$ from random play. This is the architecture's \emph{capacity}, independent of the optimizer.
The on-policy probe mixes ``the representation tracks state'' with ``the agent keeps the automaton
in a small set of states,'' so we report both. Throughout, ``recovers the state'' means linearly
recoverable from the policy's features, and nothing more. Figure~\ref{fig:instrument} summarizes
the three read-outs and the verdict they combine into.

\section{Experiments}

We measure oracle-normalized reward and the state probe on the same runs (Figure~\ref{fig:axes}). We
exclude tasks that random play already solves. A supervised test confirms that a GRU
\citep{cho2014gru} can represent the hidden state (\S\ref{sec:capacity}). The remaining experiments
vary the optimizer (\S\ref{sec:opt}), the observations and prediction losses (\S\ref{sec:aux}), and
the counter size (\S\ref{sec:scale}). We also test whether a policy encodes the state yet controls
poorly (\S\ref{sec:gaps}). Held-out automata and automata mined from checksums and process logs test the
permutation-structure warning signal (\S\ref{sec:predictor}--\S\ref{sec:mined}).

\textbf{Protocol.} Unless a section states otherwise, every RL run uses the following settings.
Episode lengths are drawn uniformly from $4$--$32$ during training. Evaluation uses lengths
$4$--$32$ (validation), $33$--$64$, and $65$--$128$ (the OOD test), all at
$p_{\mathrm{control}}{=}0.5$. The GRU policy has hidden size $128$ (about $63$k parameters).
Batched PPO runs $6{,}000$ updates of $64$ episodes each per seed, about $384$k episodes or
$6.9$M environment steps (the optimizer comparison of Table~\ref{tab:opt} and the $5\times$-budget
A2C runs use $32$ episodes per update). The optimizer is Adam on a cosine schedule from
$3{\times}10^{-4}$ to $3{\times}10^{-5}$ with $200$ warmup updates, $4$ PPO epochs per batch,
clip $0.2$, minibatch $64$, $\gamma{=}0.99$, GAE $\lambda{=}0.95$. Probes read the final
checkpoint of the fixed budget, with no checkpoint selection. Probe data are fresh rollouts at
lengths $16/32/64$ with $60$--$150$ episodes per length, split $70/30$ by episode so no episode
spans the train/test boundary. Every ``$\pm$'' is a spread over seeds around the mean, with the
estimator stated in each table or figure caption. An inline $\pm$ uses the estimator of the table
it quotes ($95\%$ bootstrap confidence intervals in \S\ref{sec:opt} and \S\ref{sec:scale},
population s.d.\ in the training curves and the causal panel).

\subsection{Diagnostic tasks and capacity}\label{sec:capacity}
We keep a task only if random play sits below the oracle, and we treat a later GRU shortfall as
a learning failure only if a supervised GRU can represent the hidden state.
Table~\ref{tab:tasks} lists the tasks named in the rest of the paper.

\begin{table}[h!]\centering\footnotesize
\caption{Tasks named below.}
\label{tab:tasks}
\begin{tabular}{@{}ll@{}}
\toprule
task & accept if\\
\midrule
\texttt{contains\_0010} & contains the substring 0010\\
\texttt{parity} & even number of 1s\\
\texttt{mod\_2} & last bit is 0\\
\modk{} ($k{=}3,5,7,9,11$) & binary value is $0$ mod $k$\\
\texttt{no\_three\_zeros} & no three consecutive 0s\\
\texttt{ends\_with\_101} & ends with 101\\
\texttt{ends\_with\_1000} & ends with 1000\\
\texttt{workflow\_api} & finish an auth--fetch--submit path\\
\texttt{workflow\_longrange} & latch a credential and submit it\\
\texttt{workflow\_reconcile} & running balance is $0$ at the end\\
\texttt{mined\_div7} & binary integer is divisible by 7\\
\texttt{mined\_ean} & EAN checksum is valid\\
\bottomrule
\end{tabular}
\end{table}

The exact oracle (\S\ref{sec:testbed}) separates diagnostic tasks from trivial ones.
\texttt{contains\_0010} is non-diagnostic: random play already reaches optimal.
On \texttt{parity} and \modk{} the optimal return remains flat with horizon, so declining agent
return reflects a growing gap to the oracle.

Under supervision, a full-prefix GRU predicts $q_t$ at accuracy $1.00$, including at four times
the training length (our out-of-distribution, or OOD, test). A window-MLP is window-bounded (chance OOD).
A plain transformer fails to length-generalize parity ($0.54$, chance) despite full context,
in line with transformer-counting results (\S\ref{sec:related}). Whether a supervised model can
represent the state depends on the architecture and on whether it sees a window or the full
prefix. A full-prefix GRU can, so later GRU shortfalls are learning failures.

\subsection{Weak RL stays at random without state}\label{sec:opt}
Under deliberately weak on-policy RL (A2C, i.e.\ REINFORCE with a value baseline
\citep{mnih2016a3c}), the probe is at
chance for every architecture, including the GRU, whose supervised capacity saturates on these
tasks. On \texttt{parity}, A2C's val$_n$ matches random play ($0.66$ against a random baseline of
$0.67$). On \texttt{mod\_3} with broadcast credit the same holds ($0.51$ against $0.48$). Reward
and probe performance do not always move together. Later experiments change them independently,
although additional training raises both on \texttt{mod\_3}.
A pre-specified control (Appendix~\ref{app:prereg}) tests whether that joint failure is an optimizer
artifact or a property of sparse terminal reward. It crosses recurrent
PPO$+$GAE$+$advantage normalization \citep{schulman2017ppo,schulman2016gae} vs.\ A2C with two
credit schemes, broadcast ($\gamma{=}1$:
every step credited with the terminal return) vs.\ discounted return-to-go, on
$\{$window-MLP, GRU$\}\times\{$parity, mod\_3$\}$, $p_{\mathrm{control}}{=}0.5$, 5 seeds.
Table~\ref{tab:opt} reports the four cells. Switching A2C to PPO, with the credit scheme held
fixed, moves the probe gap far more than switching credit schemes with the optimizer held fixed.
The gap is largest on parity. Credit still matters: on mod\_3, discounting lifts A2C, and under
PPO, broadcast edges out discounted return-to-go. Even with PPO, recovery is partial and
high-variance, well below the supervised-capacity ceiling of ${\approx}{+}0.46$/${+}0.60$.
The recurrence advantage appears in RL only under PPO (GRU$+$PPO length-generalizes mod\_3 OOD
$0.75{\pm}.20$; window-MLP $0.55{\pm}.02$; transformer $0.49{\pm}.04$).

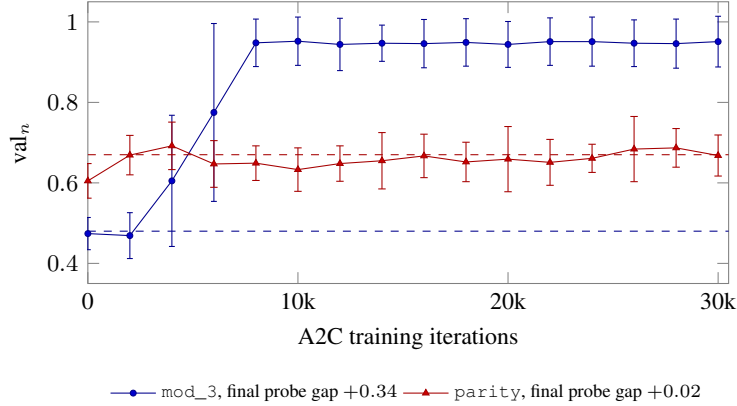
\begin{figure}[h!]\centering
% Coordinates generated by scripts/fig_a2c_budget.py from runs/a2c_long.
\begin{tikzpicture}
\begin{axis}[width=\twocol{\linewidth}{0.72\linewidth}, height=\twocol{0.55\linewidth}{0.38\linewidth},
  xlabel={A2C training iterations}, ylabel={val$_n$},
  xmin=0, xmax=30500, ymin=0.35, ymax=1.05,
  scaled x ticks=false, xtick={0,10000,20000,30000}, xticklabels={0,10k,20k,30k},
  tick label style={font=\footnotesize}, label style={font=\footnotesize}, axis line style={gray},
  legend style={font=\scriptsize, at={(0.5,-0.32)}, anchor=north, draw=none, fill=none, legend columns=2}]
\addplot+[color=blue!55!black, mark=*, mark size=1.1pt, error bars/.cd, y dir=both, y explicit] coordinates {(1,0.474) +- (0,0.040) (2000,0.469) +- (0,0.057) (4000,0.605) +- (0,0.163) (6000,0.775) +- (0,0.221) (8000,0.948) +- (0,0.059) (10000,0.952) +- (0,0.060) (12000,0.944) +- (0,0.065) (14000,0.947) +- (0,0.045) (16000,0.946) +- (0,0.060) (18000,0.949) +- (0,0.059) (20000,0.944) +- (0,0.057) (22000,0.951) +- (0,0.059) (24000,0.951) +- (0,0.061) (26000,0.947) +- (0,0.058) (28000,0.946) +- (0,0.061) (30000,0.951) +- (0,0.063)};
\addlegendentry{\texttt{mod\_3}, final probe gap $+0.34$}
\addplot+[color=red!65!black, mark=triangle*, mark size=1.4pt, error bars/.cd, y dir=both, y explicit] coordinates {(1,0.605) +- (0,0.043) (2000,0.669) +- (0,0.049) (4000,0.692) +- (0,0.059) (6000,0.647) +- (0,0.058) (8000,0.649) +- (0,0.043) (10000,0.633) +- (0,0.054) (12000,0.648) +- (0,0.044) (14000,0.655) +- (0,0.070) (16000,0.667) +- (0,0.054) (18000,0.652) +- (0,0.049) (20000,0.659) +- (0,0.081) (22000,0.651) +- (0,0.057) (24000,0.661) +- (0,0.035) (26000,0.684) +- (0,0.081) (28000,0.687) +- (0,0.048) (30000,0.668) +- (0,0.051)};
\addlegendentry{\texttt{parity}, final probe gap $+0.02$}
\draw[red!65!black, dashed] (axis cs:0,0.67) -- (axis cs:30500,0.67);
\draw[blue!55!black, dashed] (axis cs:0,0.48) -- (axis cs:30500,0.48);
\end{axis}
\end{tikzpicture}
\Description{Two training curves of oracle-normalized success against A2C iterations: mod_3 climbs
from its random-play baseline to near one by 8000 iterations, while parity stays flat at its
random-play baseline for all 30000 iterations.}%
\caption{A2C at $5\times$ the budget (GRU, $10$ seeds, mean${\pm}$s.d.). Dashed lines are
random play ($0.48$ on \texttt{mod\_3}, $0.67$ on \texttt{parity}). \texttt{mod\_3} escapes and
its probe rises with reward; \texttt{parity} stays locked at random with a chance probe.}
\label{fig:budget}
\end{figure}

\textbf{More budget does not always escape.} One worry is that A2C was under-run.
Running it for $5\times$ the budget ($30$k iters, $10$ seeds) gives different results across tasks
(Figure~\ref{fig:budget}). On \texttt{mod\_3} the extra budget raises reward and the probe together. val$_n$ climbs
$0.51\!\to\!0.95$
and the probe $+0.02\!\to\!+0.34$, near the $+0.36$ probe gap that the privileged state
auxiliary of \S\ref{sec:aux} reaches. The chance probe was
under-training. On \texttt{parity} A2C stays locked: val$_n$ is flat at $0.67$, matching random
play, and the probe stays at chance ($+0.02$) all the way to $30$k. More of the same optimizer
does not escape within the tested budgets. PPO escapes faster at a fixed budget
(Table~\ref{tab:opt}). Task structure
governs whether extra budget helps at all. \texttt{parity}'s chance-probe lock is consistent with
the credit bottleneck in Appendix~\ref{app:credit}, a candidate local mechanism. The bound proves
no impossibility, and we have not shown that trained policies satisfy its assumptions.
The lock holds under the tested optimizers and budgets. Oracle-potential shaping
\citep{ng1999shaping} is a change to the reward signal, not the optimizer
($F{=}\gamma\Phi(q_{t+1}){-}\Phi(q_t)$, $\Phi$ the oracle value, 5 seeds). It reaches $+0.30$
on parity and $+0.35$ on mod\_3, near the probe gaps from a loss that predicts the true state
($+0.28$/$+0.36$).
On a harder counter the recovered fraction stays well below the privileged-auxiliary reference at
a shared budget (\S\ref{sec:scale}: PPO$/$aux shrinks to $0.31$ at $k{=}11$).

\begin{widetable}[t]
\caption{Optimizer$\times$credit-scheme $2\times2$ (GRU, 5 seeds, mean$\pm$95\% bootstrap confidence interval).
Random-play val$_n$ is $0.67$ on \texttt{parity} and $0.48$ on \texttt{mod\_3}.}
\label{tab:opt}
\begin{tabular}{lcccc}
\toprule
optimizer $\times$ credit & parity val$_n$ & parity probe gap & mod\_3 val$_n$ & mod\_3 probe gap\\
\midrule
A2C $\times$ broadcast & $0.66{\pm}.04$ & $+0.01{\pm}.00$ & $0.51{\pm}.02$ & $+0.02{\pm}.01$\\
A2C $\times$ discounted & $0.65{\pm}.07$ & $+0.02{\pm}.01$ & $0.66{\pm}.18$ & $+0.15{\pm}.12$\\
PPO $\times$ broadcast & $0.79{\pm}.14$ & $+0.13{\pm}.09$ & $0.96{\pm}.08$ & $+0.34{\pm}.01$\\
PPO $\times$ discounted & $0.80{\pm}.11$ & $+0.20{\pm}.11$ & $0.80{\pm}.18$ & $+0.21{\pm}.12$\\
\bottomrule
\end{tabular}
\end{widetable}

\subsection{Auxiliary losses and observability}\label{sec:aux}
Under GRU$+$PPO (5 seeds) a \emph{privileged} state auxiliary (predict true $q_t$; an upper bound,
not deployable) adds almost nothing at low complexity, because PPO already recovers the state.
Its benefit grows where the base learner fails: under weak A2C, and at high complexity
(\S\ref{sec:scale}). We test two \emph{label-free} prediction losses. The first predicts the
alphabet symbol emitted at the current step from the
features that choose the action. The environment's draws are uniform and state-independent, but
the agent's own symbols can depend on the state through the policy, so this loss is not
structurally state-free. On \texttt{parity} and \texttt{mod\_3} it adds nothing beyond the base
learner and can lower reward. The second targets an extra observation channel that varies how much
state the observations reveal. At every step the channel shows one extra symbol. With probability
$\beta$ that symbol identifies the state reached by the step's transition. Otherwise it is
uniform noise from the same vocabulary. The auxiliary head predicts each step's channel symbol
before it is observed. At $\beta{=}0$ that target is pure noise and carries no state signal by
construction, and the probe reflects sparse PPO alone ($+0.14$ on \texttt{mod\_5};
Table~\ref{tab:panel}, Appendix~\ref{app:validation}). That isolates the case where state is
revealed only through outcome reward. On \texttt{mod\_5}, where sparse PPO alone struggles, the
channel auxiliary is fully effective by $\beta{=}0.10$, the smallest nonzero level tested (probe
gap $+0.47$, reward $0.93$, 5 seeds). An observation-only control isolates the auxiliary's share.
At $\beta{=}0.10$ with the auxiliary removed, on five matched seeds at the same budget, the
observations alone reach reward $0.90$ and probe gap $+0.38$, against $0.93$ and $+0.47$ with the
auxiliary (Table~\ref{tab:panel}). The prediction loss adds the remaining probe gap, cuts seed
variance, and lifts OOD reward from
$0.61{\pm}.03$ to $0.80{\pm}.14$. On the tested grid the rise is steep.

\subsection{Scaling and the easy/hard boundary}\label{sec:scale}
Sweeping the \modk{} family, with supervised capacity perfect at every $k$, shows two trends
(Table~\ref{tab:scale}). Sparse-reward PPO recovers most of the privileged-auxiliary probe gap at
$k{=}2,3$ and about a third at $k{=}11$. The privileged auxiliary's reward benefit grows from
about zero to its largest values. As tasks become more stateful, sparse reward recovers a
shrinking share of the recoverable state, and a state signal becomes more useful.

\begin{table}[h!]\centering\small
\caption{Scaling the \modk{} counter (GRU$+$PPO, 5 seeds, mean$\pm$95\% bootstrap confidence
interval). PPO$/$aux $=$ (PPO probe gap)/(privileged-auxiliary probe gap), the fraction of that
reference recovered by sparse reward.}
\label{tab:scale}
\twocol{\setlength{\tabcolsep}{3pt}}{}%
\begin{tabular}{ccccc}
\toprule
$k$ & PPO val$_n$ & PPO probe gap & aux$-$PPO reward & PPO$/$aux\\
\midrule
2  & $0.97{\pm}.03$ & $+0.26{\pm}.01$ & $-0.01$ & $0.83$\\
3  & $0.94{\pm}.02$ & $+0.36{\pm}.03$ & $+0.03$ & $0.99$\\
5  & $0.72{\pm}.25$ & $+0.24{\pm}.16$ & $+0.26$ & $0.52$\\
7  & $0.65{\pm}.27$ & $+0.30{\pm}.16$ & $+0.36$ & $0.61$\\
9  & $0.75{\pm}.17$ & $+0.33{\pm}.09$ & $+0.25$ & $0.55$\\
11 & $0.46{\pm}.13$ & $+0.19{\pm}.07$ & $+0.43$ & $0.31$\\
\bottomrule
\end{tabular}
\end{table}

\subsection{Perception versus planning gaps}\label{sec:gaps}
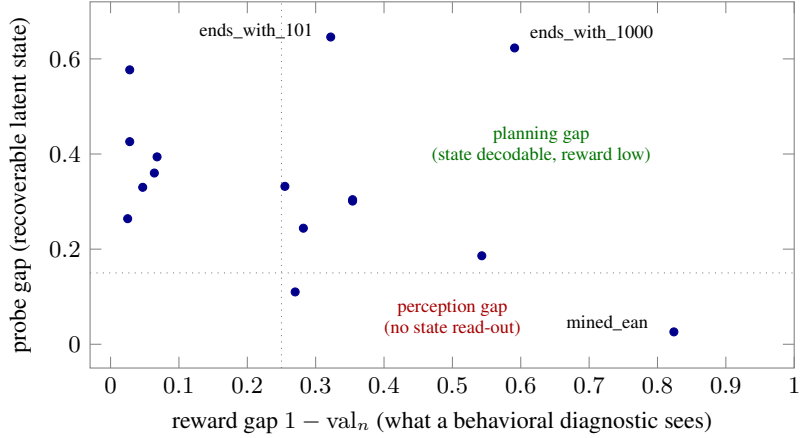
\begin{figure}[h!]\centering
\begin{tikzpicture}
\begin{axis}[width=\twocol{\linewidth}{0.78\linewidth}, height=\twocol{0.62\linewidth}{0.46\linewidth},
  xlabel={reward gap $1-\mathrm{val}_n$ (what a behavioral diagnostic sees)},
  ylabel={probe gap (recoverable latent state)},
  xmin=-0.03, xmax=1.0, ymin=-0.05, ymax=0.72,
  tick label style={font=\footnotesize}, label style={font=\footnotesize}, axis line style={gray}]
\addplot[only marks, mark=*, mark size=1.5pt, color=blue!55!black] coordinates {
 (0.270,0.110) (0.025,0.264) (0.064,0.360) (0.282,0.244) (0.354,0.304)
 (0.255,0.332) (0.543,0.186) (0.028,0.426) (0.322,0.646) (0.591,0.623)
 (0.047,0.330) (0.028,0.577) (0.068,0.394) (0.354,0.301) (0.824,0.026)};
\draw[gray,dotted] (axis cs:-0.03,0.15) -- (axis cs:1.0,0.15);
\draw[gray,dotted] (axis cs:0.25,-0.05) -- (axis cs:0.25,0.72);
\node[font=\scriptsize,anchor=west] at (axis cs:0.60,0.655) {ends\_with\_1000};
\node[font=\scriptsize,anchor=east] at (axis cs:0.31,0.66) {ends\_with\_101};
\node[font=\scriptsize,anchor=east] at (axis cs:0.80,0.045) {mined\_ean};
\node[font=\scriptsize,color=green!45!black,align=center] at (axis cs:0.63,0.42) {planning gap\\(state decodable, reward low)};
\node[font=\scriptsize,color=red!65!black,align=center] at (axis cs:0.50,0.055) {perception gap\\(no state read-out)};
\end{axis}
\end{tikzpicture}
\Description{Scatter plot of reward gap against probe gap for all RL-evaluated tasks; tasks with
the same reward gap span the full probe range, from perception gaps at the bottom to planning gaps
at the top.}%
\caption{Each point is an RL-evaluated task; at a given large reward gap the probe ranges from
chance (perception gap) to high (planning gap).}
\label{fig:behav}
\end{figure}

The results show two failure modes. \texttt{ends\_with\_1000} has a
planning gap: the probe is high ($+0.62$) yet reward is only $0.41$. The state is decodable and
control is still far from optimal. \texttt{mined\_ean} instead has a perception gap
(reward gap $0.82$, probe at chance $+0.03$, capacity saturated; \S\ref{sec:mined}).
Whether a planning-gap policy consults the decoded state is a separate question. On a second
planning task, \texttt{ends\_with\_101}, patching the decoded state direction moves the greedy
action to the $\delta$-predicted one in $0.04$--$0.47$ of cases across five seeds, against
$0.04$--$0.07$ for a norm-matched random direction. Two seeds sit at the random level and three
above it. The planning-gap label therefore means suboptimal control despite a decodable state. It
does not by itself establish that the state goes unused.
A reward-only auditor sees the oracle-normalized reward gap and the OOD reward drop. Across the
suite those signals are nearly uncorrelated with the recoverable state (Spearman with the probe
gap $-0.29$ and $-0.25$). Among the nine tasks where the agent is visibly suboptimal (reward gap
${\ge}0.25$), the behavioral signature is nearly the same, yet the probe spans $0.03$
(\texttt{mined\_ean}, perception) to $0.65$ (\texttt{ends\_with\_101}, planning). The dotted lines
in Figure~\ref{fig:behav} mark the two cuts: reward gap $0.25$, and probe gap $0.15$, below which
the read-out is at or near chance. Behavior cannot distinguish a missing read-out from a decodable
state with failing control. The two failures suggest different interventions, though
representation learning and policy optimization can each help both
(Figure~\ref{fig:behav}).

\subsection{A group-language warning signal}\label{sec:predictor}
We authored workflow shapes that each isolate a candidate explanation for
difficulty. Sparse PPO solves all of them: reach-commit API stages (steer into an absorbing
accept and stay); a long-range latch (a bit set once and held, so long-range distance is
not the problem); a threshold quota; and a terminal-invariant reconciliation (additive, with a
hold action, hence steerable to zero). The large reward-vs-state gap and auxiliary benefit
appear only in hard-to-steer counters. State count, long-range distance, and
terminal-invariance alone do not predict difficulty. Difficulty tracks terminal-exact
accumulation under poorly controllable (mixing) dynamics.

That distinction has an algebraic form computable from $(\delta,p_{\mathrm{control}})$ before any
training. Two automata with the same number of states can differ in a way that is obvious from
the transition table. On \texttt{parity} both symbols are permutations: $1$ flips even and odd, $0$
preserves them, and no two histories collapse, so the agent has to keep a running count. On
\texttt{mod\_2} the same two-state machine
is just the last bit: one symbol resets the state, and recent context is enough.
Formally, a permutation automaton is one where every symbol induces a
bijection $q\mapsto\delta(q,a)$ on the reachable states. Then no input ever merges states, the latent state never
forgets, and the task recognizes a \emph{group language} \citep{schutzenberger1965,krohn1965}.
When some symbol is non-injective, histories can merge. Across our
non-permutation examples this makes the state a function of recent context, which a recurrent
policy can track locally. That split is parameter-free and matches the $k{=}2$ vs.\ $k{\ge}3$
result: \texttt{mod\_2}$\equiv$last bit (a merging reset, easy), while \modk{} with $k\ge3$ (odd
$k$, where doubling is invertible) is a group counter that must accumulate. State count does not
decide hardness (\texttt{parity}, 2 states, is hard; \texttt{no\_three\_zeros}, 4 states, is
easy). In the curated suite, permutation structure identifies tasks that must retain information
throughout the sequence. A low-$k$ counter like \texttt{mod\_3} is a permutation automaton and is still learnable,
because partial control buys enough leverage.
Group structure alone is not enough. \texttt{workflow\_reconcile} is group-like and learnable
because it has a \emph{controllable hold}: an identity transition that freezes the state once
steered to accept.\footnote{A control-theoretic identity action, distinct from the
group-theoretic sense in which \citet{cot2025} use related language for chain-of-thought
length generalization.} A scalar proxy is the \emph{control leverage}
$L=(V_{\mathrm{opt}}-V_{\mathrm{random}})/(V_{\mathrm{full}}-V_{\mathrm{random}})$, read from the
oracle. The in-sample predictor combined both: group/permutation structure and low control leverage.

\textbf{In-sample, the predictor is exact.} Treating it as a training-free scorer ($s=L$ for
permutation automata, $s={+}\infty$ otherwise) against the measured labels (hard
$\Leftrightarrow$ privileged-aux reward benefit ${\ge}0.20$), it perfectly ranks the 15
catalog tasks (ROC-AUC $1.00$; every hard task $L\le0.50$, every easy one $L\ge0.519$). It survives
leave-one-out cross-validation $14/15$ (lone error \texttt{workflow\_reconcile} at the boundary), and
within the designed group tasks (excluding the two automata mined from real code, \S\ref{sec:mined}) it tracks hardness monotonically (Spearman$(L,\text{aux benefit})=-0.83$). A
candidate mechanism is a credit bottleneck (Appendix~\ref{app:credit}). On parity, one later random symbol
makes even and odd equally likely at the end, so the terminal reward no longer depends on an early
action. Group counters behave the same way: no symbol merges states, so the early action is still
in the hidden state, but random mixing hides it from the reward (Figure~\ref{fig:creditmix}).

\begin{figure}[h!]\centering
\resizebox{\twocol{\linewidth}{0.75\linewidth}}{!}{%
\begin{tikzpicture}[font=\small,
  st/.style={draw, rounded corners=1pt, minimum height=5.8mm, minimum width=7.2mm,
    inner sep=1pt, font=\scriptsize},
  even/.style={st, fill=blue!10, draw=blue!55!black},
  mix/.style={st, fill=gray!14, draw=gray!55, minimum width=14mm},
  acc/.style={st, fill=gray!12, draw=gray!60}]
  \node[anchor=west] at (0,2.55) {\textbf{(a)} Parity: one later random symbol};
  \node[even] (e) at (2.15,1.65) {E};
  \node[even] (o) at (2.15,0.45) {O};
  \node[mix] (em) at (5.55,1.65) {E or O};
  \node[mix] (om) at (5.55,0.45) {E or O};
  \node[font=\scriptsize, anchor=east] at (1.35,1.65) {$a_t{=}0$};
  \node[font=\scriptsize, anchor=east] at (1.35,0.45) {$a_t{=}1$};
  \draw[->, gray] (e) -- node[above, font=\scriptsize, text=gray!40!black] {env $0$/$1$} (em);
  \draw[->, gray] (o) -- node[above, font=\scriptsize, text=gray!40!black] {env $0$/$1$} (om);
  \node[font=\scriptsize, text=gray!35!black, align=left, anchor=west] at (6.55,1.05)
    {same mix\\[-1pt]reward ignores $a_t$};

  \node[anchor=west] at (0,-0.45) {\textbf{(b)} Absorbing accept/reject};
  \node[acc] (a1) at (2.15,-1.25) {Acc};
  \node[acc] (a2) at (4.15,-1.25) {Acc};
  \node[acc] (a3) at (6.15,-1.25) {Acc};
  \node[acc] (r1) at (2.15,-2.35) {Rej};
  \node[acc] (r2) at (4.15,-2.35) {Rej};
  \node[acc] (r3) at (6.15,-2.35) {Rej};
  \foreach \i/\j in {a1/a2,a2/a3,r1/r2,r2/r3}{\draw[->, gray] (\i)--(\j);}
  \node[font=\scriptsize, text=gray!35!black, align=left, anchor=west] at (6.95,-1.80)
    {never mixes\\[-1pt]spread stays};
\end{tikzpicture}}
\Description{Two panels. Parity: after the agent acts the hidden state is Even or Odd, then one
random environment symbol maps both to the same Even-or-Odd mix. Absorbing accept/reject: those
two states stay apart under every symbol.}%
\caption{Parity mixing versus an absorbing accept/reject. \textbf{(a)} One later random symbol
makes even and odd equally likely. \textbf{(b)} The accept/reject difference never mixes away.}
\label{fig:creditmix}
\end{figure}

\textbf{Proposition (mixing attenuates early credit; Appendix~\ref{app:credit},
Lemma~1--Corollary~3).} \emph{Let $M$ be a permutation automaton, $P$ its uniform random-symbol
operator, and $\sigma=\|P\|_{\mathbf{1}^{\perp}}$ its one-step contraction on mean-zero value
functions; assume $\sigma<1$ and a suffix-open-loop policy: conditional on the history through
step $t$ and the control mask, the later controlled actions depend neither on $a_t$ nor on the
symbols emitted after $t$. If step $t$ is controlled, each of the $s$ later steps is controlled
independently with probability $c$, and the score norm is bounded by $B$, then the expected
broadcast score-function credit at $t$ is at most $B\sqrt{|Q|}\,\big(c+(1-c)\,\sigma\big)^{s}$.
For parity $\sigma{=}0$, so the bound decays as $c^{s}$.}

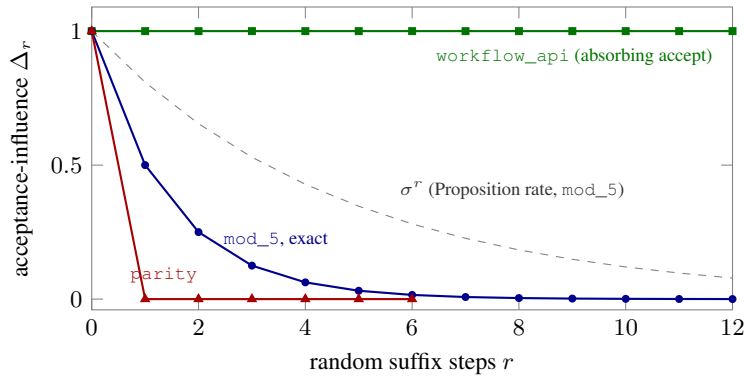
\begin{figure}[h!]\centering
% Coordinates generated by scripts/fig_credit_decay.py (exact, oracle-only).
\begin{tikzpicture}
\begin{axis}[width=\twocol{\linewidth}{0.72\linewidth}, height=\twocol{0.6\linewidth}{0.4\linewidth},
  xlabel={random suffix steps $r$},
  ylabel={acceptance-influence $\Delta_r$},
  xmin=0, xmax=12, ymin=-0.04, ymax=1.09,
  tick label style={font=\footnotesize}, label style={font=\footnotesize}, axis line style={gray}]
\addplot[color=green!45!black, thick, mark=square*, mark size=1.1pt] coordinates {
 (0,1.0000) (1,1.0000) (2,1.0000) (3,1.0000) (4,1.0000) (5,1.0000) (6,1.0000)
 (7,1.0000) (8,1.0000) (9,1.0000) (10,1.0000) (11,1.0000) (12,1.0000)};
\addplot[color=gray, dashed] coordinates {
 (0,1.0000) (1,0.8090) (2,0.6545) (3,0.5295) (4,0.4284) (5,0.3466) (6,0.2804)
 (7,0.2268) (8,0.1835) (9,0.1485) (10,0.1201) (11,0.0972) (12,0.0786)};
\addplot[color=blue!55!black, thick, mark=*, mark size=1.1pt] coordinates {
 (0,1.0000) (1,0.5000) (2,0.2500) (3,0.1250) (4,0.0625) (5,0.0312) (6,0.0156)
 (7,0.0078) (8,0.0039) (9,0.0020) (10,0.0010) (11,0.0005) (12,0.0002)};
\addplot[color=red!65!black, thick, mark=triangle*, mark size=1.5pt] coordinates {
 (0,1.0000) (1,0.0000) (2,0.0000) (3,0.0000) (4,0.0000) (5,0.0000) (6,0.0000)};
\node[font=\scriptsize, color=green!45!black, anchor=north east] at (axis cs:11.8,0.97)
  {\texttt{workflow\_api} (absorbing accept)};
\node[font=\scriptsize, color=gray!40!black, anchor=south west] at (axis cs:5.6,0.34)
  {$\sigma^{r}$ (Proposition rate, \texttt{mod\_5})};
\node[font=\scriptsize, color=blue!55!black, anchor=south west] at (axis cs:2.3,0.16)
  {\texttt{mod\_5}, exact};
\node[font=\scriptsize, color=red!65!black, anchor=south west] at (axis cs:0.55,0.03)
  {\texttt{parity}};
\end{axis}
\end{tikzpicture}
\Description{Line plot of the acceptance-influence spread against the number of random suffix
steps: parity drops to zero after one step, mod_5 halves each step and stays below the sigma-power
rate, and the absorbing workflow_api task stays at one.}%
\caption{Acceptance-influence $\Delta_r$ after $r$ random suffix steps.}
\label{fig:creditdecay}
\end{figure}

Controlled steps apply mixtures of permutation matrices, which never
expand differences among mean-zero value functions, while each uncontrolled random step contracts
those differences by $\sigma$. Averaging $\sigma^{K}$ over the binomial control mask gives the
stated rate.
The bound is on the credit of an early action when later actions ignore both it and the emitted
suffix. It does not describe how the policy learns, and a policy that reacts to later symbols can
evade it (Appendix~\ref{app:credit}).
Figure~\ref{fig:creditdecay} plots that credit as the acceptance-influence $\Delta_r$
after $r$ random suffix steps. On \modk{} it decays geometrically, below the $\sigma^{r}$
rate. Parity collapses in one step, and an absorbing accept never mixes away. Merging structure
escapes by forgetting; a controllable hold raises leverage.

\begin{widefigure}[t]
\begin{minipage}[c]{0.46\linewidth}\centering
\begin{tikzpicture}[font=\scriptsize, >=stealth,
  st/.style={draw, circle, minimum size=5mm, inner sep=0pt},
  zero/.style={->, blue!55!black, thick}, one/.style={->, red!65!black, thick}]
  \node[anchor=west, font=\small] at (-0.3,1.75) {\textbf{(a)}};
  \node[font=\small] at (1.35,1.45) {permutation};
  \node[st] (p0) at (0.6,0.55) {0};
  \node[st] (p1) at (2.1,0.55) {1};
  \node[st] (p2) at (1.35,-0.65) {2};
  \draw[zero] (p0) to[loop left] node[left] {0} (p0);
  \draw[one] (p0) to[bend left=18] node[above] {1} (p1);
  \draw[one] (p1) to[bend left=18] node[below] {1} (p0);
  \draw[zero] (p1) to[bend left=18] node[right] {0} (p2);
  \draw[zero] (p2) to[bend left=18] node[left] {0} (p1);
  \draw[one] (p2) to[loop below] node[below] {1} (p2);
  \node[font=\small] at (5.35,1.45) {merging};
  \node[st] (m0) at (4.6,0.55) {0};
  \node[st] (m1) at (6.1,0.55) {1};
  \draw[zero] (m0) to[loop left] node[left] {0} (m0);
  \draw[zero] (m1) to[bend left=18] node[below] {0} (m0);
  \draw[one] (m0) to[bend left=18] node[above] {1} (m1);
  \draw[one] (m1) to[loop right] node[right] {1} (m1);
  \node[font=\scriptsize, text=gray!40!black, anchor=north, align=center] at (1.35,-1.25) {\texttt{mod\_3}: every symbol\\is a bijection};
  \node[font=\scriptsize, text=gray!40!black, anchor=north, align=center] at (5.35,-1.25) {\texttt{mod\_2}: symbol 0 sends\\both states to 0};
\end{tikzpicture}
\end{minipage}\hfill
\begin{minipage}[c]{0.50\linewidth}\centering
% Coordinates generated by scripts/fig_heldout_strata.py from runs/_randval_summary.json.
\begin{tikzpicture}
\begin{axis}[xbar stacked, width=\linewidth, height=3.1cm, bar width=7pt,
  xmin=0, xmax=68, xlabel={held-out automata (of 153)},
  symbolic y coords={non-perm,perm-high-L,perm-low-L}, ytick=data,
  y dir=reverse, yticklabel style={font=\scriptsize\ttfamily},
  tick label style={font=\footnotesize}, label style={font=\footnotesize}, axis line style={gray},
  legend style={font=\scriptsize, at={(0.5,-0.62)}, anchor=north, draw=none, fill=none, legend columns=3},
  legend image code/.code={\draw[#1] (0cm,-0.08cm) rectangle (0.25cm,0.08cm);}]
\addplot[fill=red!65!black, draw=none] coordinates {(56,perm-low-L) (30,perm-high-L) (20,non-perm)};
\addplot[fill=green!45!black, draw=none] coordinates {(5,perm-low-L) (3,perm-high-L) (13,non-perm)};
\addplot[fill=gray!45, draw=none] coordinates {(4,perm-low-L) (5,perm-high-L) (17,non-perm)};
\legend{perception gap, planning gap, solved}
\end{axis}
\end{tikzpicture}
\par\smallskip{\small\textbf{(b)} outcome by stratum}
\end{minipage}
\Description{Left: a three-state permutation automaton in which every symbol is a bijection,
beside a two-state automaton in which symbol 0 merges both states. Right: stacked bars for the
three held-out strata; both permutation strata are mostly perception gaps, the non-permutation
stratum is mixed.}%
\caption{The warning signal. \textbf{(a)} The property: in a permutation automaton every symbol
permutes the states, so no history is forgotten; in a merging automaton a symbol collapses
histories. \textbf{(b)} Held-out outcomes by stratum (Table~\ref{tab:val}).}
\label{fig:strata}
\end{widefigure}

\begin{widetable}[t]
\caption{Held-out validation on 153 fresh random \dfa{}s (GRU$+$PPO, $5$ seeds per automaton;
supervised capacity saturated). \emph{perception} $=$ mean probe recovery ${<}0.5$;
\emph{planning} $=$ state recovered but val$_n{<}0.85$.}
\label{tab:val}
\begin{tabular}{lccccc}
\toprule
stratum (predicted) & $n$ & perception & planning & solved & mean probe gap\\
\midrule
\texttt{perm-low-L} (predicted hard)  & 65 & 56 & 5 & 4 & $+0.13$\\
\texttt{perm-high-L} (predicted easy) & 38 & 30 & 3 & 5 & $+0.16$\\
\texttt{non-perm} (predicted easy)    & 50 & 20 & 13 & 17 & $+0.35$\\
\midrule
\textbf{permutation} & 103 & 86 & 8 & 9 & ---\\
\textbf{non-permutation} & 50 & 20 & 13 & 17 & ---\\
\bottomrule
\end{tabular}
\end{widetable}

\textbf{Held-out, permutation structure predicts the perception gap.} To test generalization
beyond the curated suite, we ran the full instrument on 153 fresh
random \dfa{}s, stratified to populate both predictor classes, predicted from $\delta$ before any
RL (GRU$+$PPO, $5$ seeds per automaton; Appendix~\ref{app:validation}).
Each fresh automaton has $3$--$7$ states and $2$--$3$ symbols. In the permutation strata each
symbol's transition is a uniformly random permutation of the states; otherwise every transition
entry is drawn uniformly at random. The accept set is a random subset of up to half the states.
An automaton is kept only if it is diagnostic at $T{=}32$. The oracle must beat random play by
more than $0.08$, full control must beat random play, and the optimum must stay below $0.98$.
A perception gap here means the network could represent the state (off-policy supervised probes
saturate) but the trained agent does not reach the state-recovery threshold. An automaton gets that label if mean
probe recovery across its seeds is below $0.5$.
Of the $103$ permutation automata, $86$ have perception gaps (precision $0.83$, $95\%$ Wilson
CI $[0.75,0.89]$); of the $106$ perception gaps, $86$ are permutation (recall $0.81$).
Figure~\ref{fig:strata} shows the property and the outcome by stratum. A
$2{\times}2$ test of permutation versus perception gap gives odds ratio $7.6$ (Fisher's exact
$p{<}10^{-6}$). The suite is stratified, so these rates
depend on the mixture (Table~\ref{tab:val}). Permutation structure predicts perception gaps with
high precision on these held-out automata. The converse does not hold: $20$ of the $50$
non-permutation automata also have perception gaps. The leverage threshold does not transfer: $L$
over-credits steerability, so we keep it only as a within-group severity proxy. The original $33$
reproduce ($19/23$), and growing the set to $153$ leaves precision unchanged. An LSTM replica on
those $33$ agrees (precision and recall $0.78$). The classification is not an artifact of the
linear read-out or of on-policy visitation (Table~\ref{tab:readout}). An MLP probe keeps $103$ of
the $106$ gaps at unchanged precision. Probing the frozen encoders on uniform random play removes
the agent's visitation skew and sharpens the signal: precision rises to $0.90$ under both
read-outs, with on-policy state coverage already ${\ge}0.93$ (Appendix~\ref{app:validation}).
The classification also does not hinge on the $0.5$ recovery threshold. Sweeping it from $0.3$
to $0.7$ moves precision between $0.75$ and $0.87$ and recall between $0.90$ and $0.78$, with no
value at which the signal degrades (Appendix~\ref{app:validation}). The automata are sampled, not
minimized. As a control-based coarsening, we merge states that share the optimal value and the
optimal-action set at every remaining horizon and re-score recovery against that quotient. No
classification changes (Appendix~\ref{app:validation}). The coarsening can still keep distinctions
that an optimal action at the current horizon does not need, so it is a robustness check and does
not establish a minimal necessary representation.

\begin{table}[h!]\centering\small
\caption{Perception-gap classification under four read-outs ($153$ automata, $5$ seeds);
``kept'': the $106$ main gaps that remain.}
\label{tab:readout}
% Body generated by scripts/gen_probe_ext_table.py from runs/_probe_ext_summary.json.
\begin{tabular}{lcccc}
\toprule
read-out & gaps & kept & precision & recall\\
\midrule
linear, on-policy (main) & 106 & 106 & $0.83$ & $0.81$\\
MLP, on-policy & 103 & 103 & $0.83$ & $0.83$\\
linear, random play & 112 & 102 & $0.90$ & $0.83$\\
MLP, random play & 109 & 100 & $0.90$ & $0.85$\\
\bottomrule
\end{tabular}
\end{table}
A val$_n$ well above random with no state is rare in this random population (median
$0.48$ across the $106$ gaps; only $2$ of $106$ reach ${\ge}0.85$).
The non-permutation perception gaps mix those rare high-reward cases with ordinary optimization
failures. Permutation structure predicts sub-threshold state recovery
under this RL procedure. It does not, by itself, predict a reward-successful shortcut.
A2C on \texttt{parity} is the other end: val$_n$ locked at random with a chance probe
(\S\ref{sec:opt}).

Where the predictor is imperfect, a controllable hold and informative observations still help
on a known-hard task (Appendix~\ref{app:validation}, Table~\ref{tab:panel}).
On \texttt{mod\_5}, adding a controllable hold (still a permutation automaton) mitigates: reward
$0.50{\to}0.89$ and the probe $+0.17{\to}+0.27$, about half the ${\approx}{+}0.47$
privileged-auxiliary probe gap, with
reduced seed variance. Informative observations with the channel auxiliary
(\S\ref{sec:aux} at $\beta{\ge}0.1$) close the gap: $0.93$ and $+0.47$. The audit uses only the
oracle and probe. On \texttt{mod\_5} only the observation change fully closes the gap.

\subsection{Instances mined from code and logs}\label{sec:mined}
To test whether the failure mode arises in structure we did not design, we mine hidden \dfa{}s
from implementations of deployed checksum rules via RPNI \citep{oncina1992,angluin1987} (passive
automaton learning) and run the same instrument. We target checksum and divisibility rules, which
are ubiquitous, self-labeling, and structurally in our hard regime. That checks the pipeline and
shows the hard structure exists in deployed rules. It does not claim the structure is common. \texttt{mined\_div7}
(binary-value-divisible-by-7, recovered exactly, capacity $1.0$) reproduces the full
\texttt{mod\_7} gap: sparse PPO is partial ($0.65{\pm}.27$, probe $+0.30$), while the privileged
auxiliary and a deployable label-free one ($\beta{=}1$) both rescue it to optimal.
\texttt{mined\_ean}, a 20-state, 10-symbol EAN-checksum automaton, is the second hard instance.
The default $10$k-sample capacity bound underfits its 10-symbol alphabet. At $104$k samples the
bound saturates, with validation and OOD accuracy $1.0$, while the trained agent's probe stays at
chance (\S\ref{sec:gaps}). Both mined automata are minimal. The
three-axis pattern appears on transition structure the miner recovered from those
implementations, not structure we designed for the suite.
All seven real process logs we mined
\citep{bpic12,bpic15,bpic17,bpic19,hospitalbilling,roadfines,sepsis} (procurement, clinical, two
loan-application, building-permit, billing, and fines processes) are non-permutation. Five mine to
2-state reach-commit automata and two are decide-late automata (the label depends only on a short
final suffix). All are learnable. These results concern the mined approximations, not the full
workflows. Real checksums are permutation automata and perception-hard, while every mined
workflow \dfa{} is non-permutation and learnable. The perception gap exists in deployed checksum
rules and is not the common case among the logs we surveyed.

\section{Discussion, Limitations, and Conclusion}
The testbed measures whether an agent learns the hidden state behind its reward. In the base
full-prefix GRU setting, state learning is gated by credit assignment and
reward design, not by the observation and not by capacity (the GRU's supervised bound is
saturated). Reward and latent-state learning are only weakly related under sparse reward. A
stronger optimizer tightens the relationship, while tasks that require more accumulation weaken it.
In the seven auxiliary-installed policies we tested interventionally, the decoded state is used:
editing the hidden activation along the decoded direction shifts the greedy action to the
$\delta$-predicted one far more than a norm-matched random direction does ($0.32$--$0.68$ vs.\
$0.00$--$0.21$). The sparse-PPO planning-gap policies are mixed. Across five
\texttt{ends\_with\_101} seeds the shift ranges from the random level to $0.47$
(\S\ref{sec:gaps}). Any
workflow represented as an explicit FSM with known transition and reward structure can be given
an exact oracle and a state probe and audited white-box.
Permutation (group-language) structure is a warning signal for the perception gap (held-out
precision $0.83$, $86$ of the $103$ flagged, capacity controlled; \S\ref{sec:predictor}).
Informative observations close the gap on \texttt{mod\_5}; a hold mitigates it. The audit is the
reusable output. All seven mined workflow \dfa{}s are learnable.

\textbf{Towards LLM agents (preliminary).} A Qwen2.5 port reproduces declining state recovery
with horizon under LoRA adaptation \citep{hu2022lora}. Across 1.5B--14B models, recovery on the
$k{\ge}3$ counters falls from $0.48$--$0.62$ at horizon $8$ to $0.13$--$0.25$ at $24$
(Appendix~\ref{app:llm}, Table~\ref{tab:llmcap}). These are achieved recoveries. They do not
bound what the architecture could represent. RL on 3B/7B reproduces the optimizer ordering on a
reach-commit control, and five of six \texttt{mod\_7} runs transiently approach oracle return
before falling back. A reward-successful perception gap was not observed. State was already
partly decodable at learnable horizons and remained so during training.

\textbf{Limitations.} The synthetic tasks have at most $11$ states and mostly binary alphabets,
while the mined instances reach $20$ states and $10$ symbols. Failures depend on optimizer and
budget. Additional A2C training helps \texttt{mod\_3} but not \texttt{parity} (\S\ref{sec:opt}).
The credit analysis assumes strictly contracting permutation dynamics and a suffix-open-loop
policy. It is a candidate mechanism and proves nothing about learning dynamics. The held-out warning
is one-directional. Non-permutation automata can also fail, and the leverage threshold does not
transfer (\S\ref{sec:predictor}). Predicting difficulty within these classes remains open. Both
hard mined instances are targeted checksums. Seven learnable mined workflow approximations do not
establish broader workflow prevalence. Causal fixes are demonstrated on \texttt{mod\_5}, with
three seeds at $\beta{=}0$ and five at $\beta{\ge}0.1$. Larger automata and richer real-world
sources remain future work.

\textbf{Conclusion.} Hidden partial-control \dfa{}s let us measure whether reward success reflects
latent-state learning. The two can diverge. Optimizer strength, task structure, and observation
informativeness help explain when they do. On the held-out automata, permutation structure is a
high-precision warning for perception gaps, although non-permutation automata can also fail. Reward
alone cannot distinguish a state the probes fail to recover from a state that is decodable while
control still fails.

\label{endmain}

\section*{Reproducibility statement}
The accompanying repository releases the environment, exact oracle, probes, supervised
capacity bound, and every training and analysis stage. \texttt{reproduce.sh} reruns the
pipeline. The README lists each stage and its cost. The script uses the fixed
153-automaton suite already in the repository rather than drawing a new one, and it
does not download the seven public process-event logs (those must be fetched from
4TU.ResearchData before the log-mining stage). The LLM appendix is a feasibility
check and is not part of the main claims, so the default script does not run it.

% ---------------- References (do NOT count toward the 9-page limit) ----------------

% ======================= APPENDICES (do not count toward page limit) =======================
\normalsize
\setcounter{section}{0}
\renewcommand{\thesection}{\Alph{section}}
% Unique hyperref destinations for the lettered appendices (the section counter values 4--8 would
% otherwise collide with main-text sections 4--6 and warn about duplicate destinations).
\renewcommand{\theHsection}{appendix.\arabic{section}}

\section{Pre-specification of the optimizer-vs-reward-channel control}\label{app:prereg}
The \S\ref{sec:opt} correction rests on a control we fixed before running it; the hypotheses,
design, and decision rule below all predate the PPO results. Under weak on-policy RL the probe sat at
chance for every architecture; we wrote down two competing explanations and a design to separate them.

\textbf{Pre-specified hypotheses.} \textbf{H$_{\mathrm{opt}}$ (optimizer):} the chance probe is an
artifact of a weak optimizer; a stronger optimizer (recurrent PPO$+$GAE$+$advantage normalization)
will raise the probe and reward together. \textbf{H$_{\mathrm{signal}}$ (reward channel):} the chance
probe is intrinsic to sparse terminal reward. Changing the credit scheme (broadcast return
vs.\ discounted return-to-go), with the optimizer held fixed, would then move the probe.

\textbf{Pre-specified design (fixed before seeing PPO results).} A $2\times2$ grid, optimizer
$\{$A2C, PPO$\}$ $\times$ credit-scheme $\{$broadcast ($\gamma{=}1$), discounted return-to-go$\}$, crossed with
$\{$window-MLP, GRU$\}\times\{$\texttt{parity}, \texttt{mod\_3}$\}$, at $p_{\mathrm{control}}{=}0.5$,
5 seeds, identical network sizes, horizon, and evaluation protocol across cells. The GRU's off-policy
supervised capacity (${=}1.0$) was established first, so any cell's shortfall is a learning effect
and not a representational one.

\textbf{Pre-specified decision rule.} Support H$_{\mathrm{opt}}$ if the PPO cells raise the probe gap
substantially over A2C holding the credit scheme fixed; support H$_{\mathrm{signal}}$ if the
discounted cells do so holding the optimizer fixed. \textbf{Outcome
(Table~\ref{tab:opt}):} the optimizer axis moved the probe (A2C${\to}$PPO at fixed credit scheme),
the credit-scheme axis was secondary, supporting H$_{\mathrm{opt}}$ and forcing us to overturn
the too-strong reading that sparse-reward RL cannot install latent state. Recovery remains
partial and seed-variable, which is the claim that survives in \S\ref{sec:opt}.

\section{Why mixing permutation automata attenuate early credit}\label{app:credit}
This appendix derives the mechanism behind the warning signal in \S\ref{sec:predictor}. The result
is a local credit-assignment analysis under explicit assumptions. It is neither an impossibility
result nor a formal statement about the learning dynamics.

\textbf{The picture, before the bound.} Take parity, where the hidden state is even or odd.
The agent acts at some step $t$, then the episode continues. If any later step is a uniformly
random environment symbol, even and odd become equally likely. The final accept/reject
no longer depends on the action at $t$, so the terminal reward is silent about that action.
A group counter behaves the same way. No symbol ever merges two states, so the hidden state still
contains the early action. Random mixing makes every state about equally likely, however, and
removes the action's effect on reward. Near a policy that has not learned to react to the stream,
the learning signal on early actions shrinks geometrically with the number of later random steps.
The bound below describes this bottleneck.
Figure~\ref{fig:creditmix} (main text) draws the mixing and the absorbing contrast.

\textbf{Setup.} An episode has horizon $T$. At each step the action is the agent's with
probability $c=p_{\mathrm{control}}$ and a uniform random symbol otherwise; $\mathcal{C}$ is the
set of controlled steps. The reward is the terminal indicator $R=\mathbf{1}[q_T\in F]$, and we use
the broadcast ($\gamma{=}1$) estimator
$\nabla_\theta J=\mathbb{E}\big[R\sum_{t\in\mathcal{C}}\nabla_\theta\log\pi_\theta(a_t\mid h_t)\big]$.
The policy conditions on the observed prefix $h_t$ only, with a bounded score,
$\mathbb{E}\,\|\nabla_\theta\log\pi_\theta(a_t\mid h_t)\|\le B$.

\textbf{Lemma 1 (credit is controlled by the action-value spread).}
\emph{Fix $h_t$ with $t\in\mathcal{C}$, write $Q_t(h_t,a)=\Pr[R{=}1\mid h_t,\,a_t{=}a]$, and let
$d_t(h_t)=\max_a Q_t(h_t,a)-\min_a Q_t(h_t,a)$. Then}
\[
\big\|\mathbb{E}[\,R\,\nabla_\theta\log\pi_\theta(a_t\mid h_t)\mid h_t\,]\big\|
\;\le\;\tfrac12\,d_t(h_t)\;
\mathbb{E}\big[\|\nabla_\theta\log\pi_\theta(a_t\mid h_t)\|\;\big|\;h_t\big].
\]
\emph{Proof.} Because $\mathbb{E}_{a_t}[\nabla_\theta\log\pi_\theta(a_t\mid h_t)]=0$, subtracting
any baseline $\bar\rho(h_t)$ from the coefficient $Q_t$ leaves the expectation unchanged.
Taking $\bar\rho$ to be the midpoint of the range of $Q_t(h_t,\cdot\,)$ bounds the coefficient
by $d_t(h_t)/2$, and the triangle inequality gives the bound. If $Q_t$ does not depend on $a_t$,
the spread is zero and the step contributes exactly zero. $\square$

The lemma is exact for any prefix-conditioned policy. The mechanism enters through $d_t$: when
does the action-value spread shrink?

\textbf{Suffix-open-loop assumption.} Conditional on $h_t$ and the control mask, the distributions
of the controlled actions after step $t$ do not depend on $a_t$ or on the symbols emitted after
$t$. A history-independent policy (a uniform initialization, for example) satisfies this. It is a
local model of a policy that has not yet learned to react to the stream; an adaptive policy can
violate it (below).

\textbf{Proposition 2 (mixing contracts the spread).}
\emph{Let $M$ be a permutation automaton, $P=\frac{1}{|\Sigma|}\sum_{a}P_a$ its uniform
random-symbol operator, and $\sigma=\|P\|_{\mathbf{1}^{\perp}}$ the factor by which one random
symbol shrinks differences among states (the second singular value of $P$ on mean-zero functions);
assume
$\sigma<1$, i.e., $P$ is a strict one-step $L_2$ contraction on mean-zero value functions (for
parity $\sigma=0$; for the doubling counter at $k{=}5$, $\sigma=\cos(\pi/5)\approx0.81$). Under the suffix-open-loop assumption, if $k$ of the
steps after $t$ are uncontrolled, then $d_t(h_t)\le 2\sqrt{|Q|}\,\sigma^{k}$.}

\emph{Proof sketch.} Given the mask, $Q_t(h_t,\cdot\,)$ is obtained from the terminal indicator by
alternately applying $P$ (uncontrolled steps) and fixed action-mixture operators (controlled
steps). Each of those operators is a convex combination of permutation matrices, so each is
doubly stochastic. Doubly stochastic operators cannot increase $L_2$ distances among mean-zero
value functions, and each application of $P$ shrinks those distances by $\sigma$. The spread of a
vector is at most twice its centered $L_2$ norm, and the
terminal indicator's centered $L_2$ norm is at most $\sqrt{|Q|}$. For parity, one uncontrolled
step makes the state uniform, so $d_t=0$ exactly. $\square$

\textbf{Corollary 3 (random control attenuates early credit exponentially).}
\emph{If each of the $s$ steps after $t$ is uncontrolled independently with probability $1-c$,
then averaging $\sigma^{K}$ over the binomial mask gives}
\[
\mathbb{E}\,[d_t]\;\le\;2\sqrt{|Q|}\,\big(c+(1-c)\,\sigma\big)^{s},
\]
\emph{so with Lemma~1, conditional on step $t$ being controlled, its expected gradient
contribution is at most $B\sqrt{|Q|}\,(c+(1-c)\sigma)^{s}$ (one further factor of $c$ if the
expectation includes the mask). For parity ($\sigma{=}0$) the surviving term is $c^{s}$, the probability that every
remaining step is controlled.}

\textbf{Scope of the bound.} Near a history-independent policy, random
transitions on a strictly contracting permutation automaton exponentially attenuate the direct
score-function credit reaching early decisions. That direct signal is not $\Theta(1)$. This is a
local bootstrap bottleneck, not an optimization fixed point and not an impossibility.
Late-step gradients can still shape shared recurrent parameters. The same mixing is why cyclic
counters sit on the hard side of the transformer counting literature:
\texttt{mod\_2} is a local last-bit task, while $k\ge 3$ must accumulate. An adaptive policy can
evade the bound by reacting to the emitted suffix. For parity, recording one random symbol $b$ and
replaying it later cancels its effect ($q_T=q_t\oplus a_t\oplus b\oplus b$). That evasion needs
only local memory of recent symbols, not the probed state representation. The chance-probe lock
of \S\ref{sec:opt} is the observed signature of this bottleneck. The escapes differ. \emph{State forgetting}:
non-permutation transitions let different histories merge; in our short-memory examples these
mergers repeatedly erase dependence on old symbols, so the control-relevant state is recoverable
from recent context. \emph{Value preservation}: an absorbing accept/reject difference
never mixes away, so the spread stays large and credit survives (\texttt{workflow\_api},
\texttt{no\_three\_zeros}). Controllable hold (empirical): a hold action does not stop
contraction at uncontrolled steps, but it raises finite-horizon leverage and allows late
correction; \texttt{workflow\_reconcile} is a permutation automaton that is learnable this way.

\textbf{Empirical check.} We compute the \emph{acceptance-influence} $\Delta$, the spread over
latent states of the acceptance probability under uniform play, exactly from the oracle. The
decay is consistent with the predicted geometric attenuation. Parity collapses to $\Delta=0$ after
a single random step, and \modk{} decays geometrically ($\Delta\approx2^{-r}$ after $r$ random
steps: $0.50,0.25,0.125,0.031,\dots$). \texttt{workflow\_api} stays at $\Delta\equiv1.0$
(absorbing accept/reject) and \texttt{no\_three\_zeros} stays high ($\Delta\approx0.30$ out to
$16$ steps). \texttt{workflow\_reconcile} decays like a counter under random play; the trained
policy uses its hold action to avoid that regime. The vanishing-credit regime aligns with the predictor's
hard class on the curated examples.

\section{Held-out validation of the predictor, and the causal fix}\label{app:validation}
\textbf{Fresh-DFA validation.} We generated fresh random \dfa{}s to test the predictor beyond the
curated suite. Each has $3$--$7$ states and $2$--$3$ symbols, and the oracle filter retains only
diagnostic tasks. We stratified the sample to populate both predictor classes. The predicted-hard
class, consisting of permutation automata with low leverage, makes up about $1\%$ of uniformly
random diagnostic \dfa{}s; an unstratified sample would therefore require thousands. We report
$153$ automata: an initial $33$ and a $120$-automaton extension from a separate seed, with $5$ seeds
per automaton. All
$153$ automata ship with the released code (\texttt{randdfa/suite.json}: transition tables,
predictor scores, and strata), with the measured labels in the released run outputs. Released data
keeps the original stratum keys (\texttt{hard}/\texttt{easy\_perm}/\texttt{easy\_synch}). The
predictor was computed from $\delta$ alone before any RL; we then ran the full instrument
(GRU$+$PPO, $6000$ iters). Capacity is controlled directly: an off-policy supervised probe
saturates (val and OOD accuracy ${\approx}1.0$) on the automata, so a low on-policy probe is a
perception gap (state recoverable but not recovered), not a representational limit. Each automaton
is labeled perception-hard if mean probe recovery across its seeds is ${<}0.5$, planning-hard if the state is
recovered but reward is suboptimal (val$_n<0.85$), else solved (Table~\ref{tab:val}, main text). The reward
threshold moves only the planning/solved split ($15/32$, $21/26$, and $31/16$ at thresholds
$0.80$, $0.85$, $0.90$, and $34/13$ under random-baselined normalization at $0.85$). The perception
count, and with it the permutation signal's precision and recall, does not depend on it.

\textbf{Recovery-threshold sensitivity.} The perception label itself rests on the $0.5$ recovery
cut, so we sweep that cut as well (Table~\ref{tab:thresh}). From $0.3$ to $0.7$ the gap count
grows from $86$ to $116$ while precision and recall move between $0.75$/$0.90$ and
$0.87$/$0.78$. The off-policy read-out is sharper at every value. No threshold degrades the
classification.

\begin{table}[h]\centering\small
\caption{The permutation signal across recovery thresholds ($153$ automata, $5$ seeds), under the
main linear on-policy read-out and the linear random-play read-out.}
\label{tab:thresh}
% Body generated by scripts/analyze_probe_threshold.py from runs/_probe_ext_summary.json.
\begin{tabular}{lcccccc}
\toprule
& \multicolumn{3}{c}{linear, on-policy} & \multicolumn{3}{c}{linear, random play}\\
threshold & gaps & precision & recall & gaps & precision & recall\\
\midrule
$0.3$ & 86 & $0.75$ & $0.90$ & 96 & $0.86$ & $0.93$\\
$0.4$ & 99 & $0.82$ & $0.85$ & 104 & $0.88$ & $0.88$\\
$0.5$ & 106 & $0.83$ & $0.81$ & 112 & $0.90$ & $0.83$\\
$0.6$ & 112 & $0.86$ & $0.80$ & 120 & $0.92$ & $0.79$\\
$0.7$ & 116 & $0.87$ & $0.78$ & 125 & $0.96$ & $0.79$\\
\bottomrule
\end{tabular}
\end{table}

\textbf{Minimality and control-relevance.} The held-out automata are sampled, not minimized. Of
the $153$, $26$ contain states unreachable from the start state, and $10$ contain
Nerode-equivalent reachable states (three have both, so $33$ are non-minimal). Unreachable states
never occur in a rollout and so never enter probe training or scoring. Mergeable states raise a
sharper worry. A probe scored on distinctions the accepted language never uses, or on distinctions
irrelevant to optimal play at the remaining horizon, could make a sound representation look
deficient. The oracle gives a direct check. We group states that share the optimal value and the
optimal-action set at every remaining horizon and probe that quotient
(\texttt{scripts/analyze\_control.py}); it is coarser than the full state on $28$ automata and
subsumes every Nerode merge. The grouping is conservative. Agreement at every horizon is
sufficient for interchangeability, not necessary, so the quotient is a control-based coarsening
and does not give a minimal necessary representation. Re-scoring all checkpoints against the quotient
reclassifies no automaton. The gap set stays at the same $106$, with precision $0.83$ and recall $0.81$. The
curated and mined automata are all minimal (\texttt{scripts/analyze\_minimality.py}).

The warning signal also holds in this set. Permutation structure predicts the perception gap with
high precision in held-out automata (precision $0.83$, $95\%$ CI $[0.75,0.89]$, recall $0.81$;
per-stratum counts in Table~\ref{tab:val}, main text). The relationship is one-directional. The
suite is stratified, so these aggregate rates
depend on the stratum mixture and are not natural-population estimates (predicted-hard automata
are ${\approx}1\%$ of uniform-random diagnostic \dfa{}s). Per-stratum precision is $56/65$ and
$30/38$. The original $33$ reproduce this ($19/23$, precision
$0.83$), and growing the set from $33$ to $153$ moves precision by less than $0.01$, so the
estimate is not small-sample. On the original $33$ the result also replicates on an LSTM ($22$ of
the $24$ GRU perception gaps shared, precision and recall $0.78$, capacity equally saturated), so
it is not a recurrence-cell artifact.

Two caveats. The converse fails: non-permutation automata are heterogeneous
(Table~\ref{tab:val}). Group structure warns that a perception gap is likely; a task without it
can still have one. No structural property we tried separates the non-group perception gaps
(diameter, minimum image size, and mixing rate do not), and the credit-vanishing mechanism largely
does not explain them either: their acceptance-influence $\Delta$ persists rather than decaying.
They are a mix of reward shortcuts (high reward, no tested probe read-out) and optimization failures. A
structural account of them is the open problem for the predictor. The leverage threshold also
fails to transfer: the two-part predictor's recall collapses because the scalar $L$ over-credits
steerability, marking most of the \texttt{perm-high-L} stratum easy although it is predominantly
perception gaps (Table~\ref{tab:val}). We therefore demote $L$ to a within-group severity proxy. A
better predictor of which permutation automata are hardest remains open.

\textbf{Read-out and distribution robustness.} Two further checks vary what the probe is and where
its data comes from (\texttt{scripts/run\_probe\_ext.sh}; per-checkpoint outputs ship with the
runs). Replacing the linear probe with a one-hidden-layer MLP leaves the classification nearly
intact: $103$ of the $106$ gaps remain gaps, and precision is unchanged ($86/103$). Probing the
frozen encoder on uniform random play removes the visitation skew of the agent's own behavior and
sharpens the signal. Under the off-policy linear probe, $102$ of the $106$ gaps remain. Ten
automata whose on-policy probes had been flattered by narrowed visitation join them, for $112$
gaps in all (precision $93/103=0.90$, recall $93/112=0.83$). The off-policy MLP gives $109$ gaps
($100$ of the $106$ remain; precision $0.90$, recall $93/109=0.85$). On-policy state coverage is
high (${\ge}0.93$ of states on average), so the skew is distributional rather than missing support.
The curated exemplars behave the same way. \texttt{mined\_ean} stays at chance under all four
read-outs. Its supervised capacity bound saturates at $104$k supervised samples, ten times the
default, with validation and OOD accuracy $1.0$. The 10-symbol alphabet needs the larger sample. The planning-gap tasks are fully decodable under all four. The partial \modk{}
recoveries shrink off-policy (\texttt{mod\_5} $0.48\to0.26$). Perception gaps are not artifacts
of read-out linearity or of on-policy visitation.

\textbf{The causal interventions are demonstrated on \texttt{mod\_5}.} Where prediction is imperfect, the two design levers still
work as interventions (Table~\ref{tab:panel}). On \texttt{mod\_5} (predicted and measured
hard), adding a controllable hold (\texttt{mod\_5\_hold}, which raises $L$ from
$0.433$ to $0.523$ and remains a permutation automaton) nearly doubles reward and roughly doubles
recovered state, and cuts seed variance. An informative emission channel with the label-free
auxiliary (\S\ref{sec:aux}'s $\beta$ lever) recovers the state and reward to near-optimal. The
hold mitigates. The observation-plus-auxiliary lever closes the gap. An observation-only control
at $\beta{=}0.1$, with the auxiliary removed on matched seeds and budget, separates the two
ingredients. The informative observations alone carry most of the effect (reward $0.90$, probe
$+0.38$). The prediction loss contributes the rest of the probe gap and matters most out of
distribution, where it lifts reward from $0.61{\pm}.03$ to $0.80{\pm}.14$ at test lengths
$65$--$128$.

\begin{table}[h]\centering\small
\caption{Causal predict$\to$fix$\to$measure panel on \texttt{mod\_5} (GRU$+$PPO).
Privileged-auxiliary reference (probe gap) $\approx +0.47$. The baseline row is the 3-seed panel replication; Table~\ref{tab:scale}'s 5-seed estimate is $+0.24$. Latchability: 3 seeds, no auxiliary. Observability:
the label-free channel auxiliary at informativeness $\beta$; the observations-alone row keeps
$\beta{=}0.1$ and removes the auxiliary, on matched seeds and budget. Cells are
mean$\pm$population s.d.\ over seeds (the two $\beta{=}0.1$ rows: 5 seeds; other rows 3).}
\label{tab:panel}
% Body generated by scripts/gen_panel_table.py from the runs/ trees it names.
\begin{tabular}{llcc}
\toprule
condition & lever & reward val$_n$ & probe gap\\
\midrule
\texttt{mod\_5} (hard baseline)            & ---                          & $0.50{\pm}.22$ & $+0.17{\pm}.16$\\
\texttt{mod\_5\_hold}                      & latchability (controllable hold) & $0.89{\pm}.09$ & $+0.27{\pm}.04$\\
\texttt{mod\_5}, $\beta{=}0$               & observability (off)          & $0.50{\pm}.24$ & $+0.14{\pm}.16$\\
\texttt{mod\_5}, $\beta{=}0.1$, no aux     & observations alone           & $0.90{\pm}.07$ & $+0.38{\pm}.02$\\
\texttt{mod\_5}, $\beta{\ge}0.1$           & observability (on)           & $0.93{\pm}.04$ & $+0.47{\pm}.02$\\
\bottomrule
\end{tabular}
\end{table}

\section{An LLM-agent instrument: a preliminary feasibility port}\label{app:llm}
The paper's motivation is stateful agent reliability, so we ported the entire instrument to a
real LLM agent. We render each step of \texttt{PartialControlDFAEnv} as natural-language tool I/O with the alphabet
relabeled to arbitrary tokens (so a pretrained model cannot pattern-match ``this is mod-$k$'')
and the \dfa{} rule kept out of the prompt. The agent emits one action token per controllable step
with no chain of thought (so an activation probe is not undercut by state offloaded into
emitted reasoning). The oracle transfers unchanged; we probe the residual stream (the hidden vector) at the action token
with the same linear probe, selectivity control, and capacity bounds. Every probe number below
uses episode-grouped train, validation, and test splits with the read-out layer chosen on the
validation episodes, as on the synthetic side. We ran this on Qwen2.5-3B/-7B-Instruct on a
single GPU (released code). The structure axis, the optimizer ordering, and the lock-and-escape
dynamics of \S\ref{sec:opt} appear on a real LLM. A reward-successful perception gap does not. The reason is a
difference in starting point between a pretrained model and a randomly initialized GRU, and we
measure it below. This is a transfer check, not a scaling result.

\textbf{The structure axis replicates as a recovery limit under the tested adaptation.} A frozen
pretrained model
already carries part of the latent state at short horizons. On \texttt{mod\_5} at $L{=}8$ the
linear probe on the frozen 3B model reads a balanced accuracy of $0.61$ against a chance level of
$0.20$, and on \texttt{parity} at $L{=}8$ it reads $0.75$ against $0.50$. At $L{=}24$ the same
probe is near chance (\texttt{mod\_5} gap $0.09$ on 3B and $0.14$ on 7B). A privileged
supervised fine-tune (LoRA, a small adapter on the base model, trained to predict $q_t$) is the
LLM analog of our capacity bound. A finite adaptation run establishes achieved recovery, a lower
bound. The architecture could hold more. The achieved recovery falls
with the horizon in the same way (Table~\ref{tab:llmcap}). Across Qwen2.5 1.5B--14B the pattern
is the same. At $L{=}8$ every counter is recoverable in part (\texttt{mod\_5} $0.52$--$0.57$
across sizes). By $L{=}24$ recovery falls to $0.13$--$0.25$ on every $k{\ge}3$ counter. Size does
not repair the decay: at $L{=}24$, 14B and 1.5B are within $0.04$ of each other on every
$k{\ge}3$ counter. The long-horizon \texttt{parity} cells are training-unstable (one LoRA run
per cell; $0.23$--$0.64$ at $L{=}24$), so we exclude \texttt{parity} from the size comparison.
Counting horizon, not state count and not model size, sets the recovery we achieve. The same
counting-horizon pattern appears in the transformer result of \S\ref{sec:opt}, here under LoRA
adaptation on a 1.5--14B transformer.

\textbf{The optimizer ordering appears on the reach-commit control; accumulation remains unsolved.}
On a partly recoverable task (\texttt{mod\_5}$@L8$, LoRA recovery $0.53$, oracle random $0.20$/optimal $0.55$),
our REINFORCE-level loop (group-normalized return, no value baseline or GAE) does not raise reward and
destabilizes (reward drifts below random as the policy leaves the base model). This is a
consistency check: it is the synthetic weak-optimizer (A2C)
regime of \S\ref{sec:opt}, which also failed to install state and which the stronger PPO$+$GAE
partially repaired. We ran the cheaper group-baseline loop first. A PPO$+$GAE optimizer for an
LLM needs a value head and GAE over the agent's decisions, which is heavier. We
then implemented that optimizer (a scalar value head, GAE over the agent's ordered
decisions, and a clipped surrogate with multiple epochs) and confirmed it works: on a reach-commit
control (\texttt{workflow\_api}) it raises reward where the REINFORCE loop stagnated. On
\texttt{mod\_5}$@L8$ it still does not learn. Over $300$ iterations and three seeds the reward
never trends upward. It ends at $0.53$ of the optimum in two seeds, near its value after the
first iteration ($0.46$--$0.57$), and at $0.27$ in the third. The state probe after training reads
gaps of $0.30$--$0.43$ over chance with selectivity $0.26$--$0.59$, about where the frozen model
started. PPO neither installs the state nor removes it. PPO recovers reach-commit and local tasks and only partially
recovers accumulation counters, and there only with ${\sim}30\times$ the updates.
At LLM scale the same optimizer learns the local task and not the accumulation
counter within budget. The optimizer-strength ordering (REINFORCE degrades; PPO is stable but
gets no traction on accumulation) and the accumulation-is-hard axis (\S\ref{sec:scale}) both
reproduce.

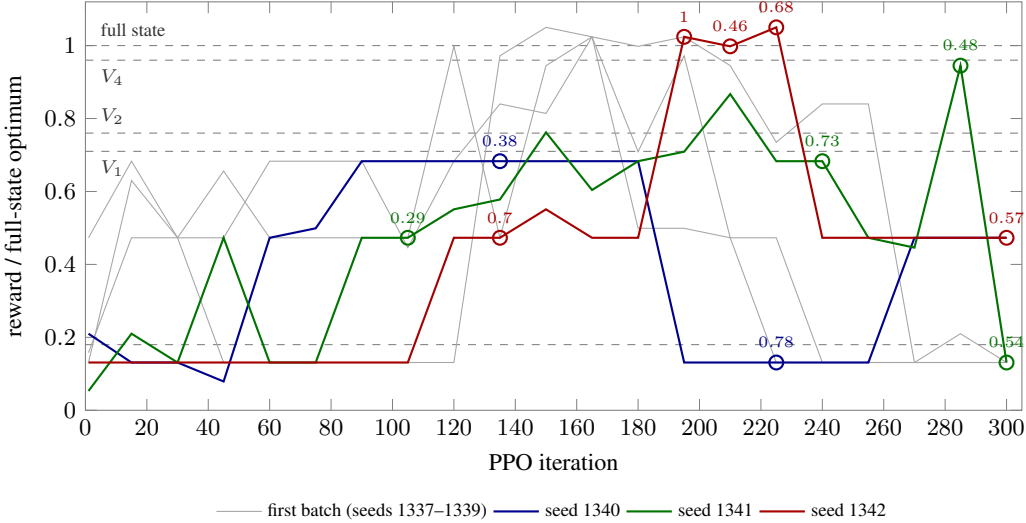
\begin{figure}[h]\centering
% Coordinates generated by scripts/fig_shortcut_arc.py from
% runs/llm/shortcut_ppo*_m7L8c8_s*/{history,adapter_probes_v2}.json.
\begin{tikzpicture}
\begin{axis}[width=\linewidth, height=0.5\linewidth,
  xlabel={PPO iteration}, ylabel={reward / full-state optimum},
  xmin=0, xmax=305, ymin=0, ymax=1.12,
  tick label style={font=\footnotesize}, label style={font=\footnotesize}, axis line style={gray},
  legend style={font=\scriptsize, at={(0.5,-0.2)}, anchor=north, draw=none, fill=none, legend columns=4},
  legend cell align=left]
\draw[gray, dashed] (axis cs:0,0.18) -- (axis cs:305,0.18);
\draw[gray, dashed] (axis cs:0,0.71) -- (axis cs:305,0.71);
\draw[gray, dashed] (axis cs:0,0.76) -- (axis cs:305,0.76);
\draw[gray, dashed] (axis cs:0,0.96) -- (axis cs:305,0.96);
\draw[gray, dashed] (axis cs:0,1.00) -- (axis cs:305,1.00);
\node[font=\scriptsize, gray!40!black, anchor=north west] at (axis cs:2,0.71) {$V_1$};
\node[font=\scriptsize, gray!40!black, anchor=south west] at (axis cs:2,0.76) {$V_2$};
\node[font=\scriptsize, gray!40!black, anchor=north west] at (axis cs:2,0.96) {$V_4$};
\node[font=\scriptsize, gray!40!black, anchor=south west] at (axis cs:2,1.00) {full state};
\addplot[gray!70, thin] coordinates {(1,0.473) (15,0.683) (30,0.473) (45,0.131) (60,0.131) (75,0.131) (90,0.131) (105,0.131) (120,0.131) (135,0.972) (150,1.050) (165,1.024) (180,0.998) (195,1.024) (210,0.945) (225,0.735) (240,0.840) (255,0.840) (270,0.131) (285,0.131) (300,0.131)};
\addlegendentry{first batch (seeds 1337--1339)}
\addplot[gray!70, thin, forget plot] coordinates {(1,0.131) (15,0.630) (30,0.473) (45,0.473) (60,0.683) (75,0.683) (90,0.683) (105,0.446) (120,0.683) (135,0.840) (150,0.814) (165,1.024) (180,0.709) (195,0.972) (210,0.473) (225,0.131) (240,0.131) (255,0.131) (270,0.131) (285,0.131) (300,0.131)};
\addplot[gray!70, thin, forget plot] coordinates {(1,0.158) (15,0.473) (30,0.473) (45,0.656) (60,0.473) (75,0.473) (90,0.473) (105,0.473) (120,0.998) (135,0.473) (150,0.945) (165,1.024) (180,0.499) (195,0.499) (210,0.473) (225,0.473) (240,0.131) (255,0.131) (270,0.131) (285,0.210) (300,0.131)};
\addplot[blue!55!black, thick] coordinates {(1,0.210) (15,0.131) (30,0.131) (45,0.079) (60,0.473) (75,0.499) (90,0.683) (105,0.683) (120,0.683) (135,0.683) (150,0.683) (165,0.683) (180,0.683) (195,0.131) (210,0.131) (225,0.131) (240,0.131) (255,0.131) (270,0.473) (285,0.473) (300,0.473)};
\addlegendentry{seed 1340}
\addplot[green!45!black, thick] coordinates {(1,0.053) (15,0.210) (30,0.131) (45,0.473) (60,0.131) (75,0.131) (90,0.473) (105,0.473) (120,0.551) (135,0.578) (150,0.762) (165,0.604) (180,0.683) (195,0.709) (210,0.867) (225,0.683) (240,0.683) (255,0.473) (270,0.446) (285,0.945) (300,0.131)};
\addlegendentry{seed 1341}
\addplot[red!65!black, thick] coordinates {(1,0.131) (15,0.131) (30,0.131) (45,0.131) (60,0.131) (75,0.131) (90,0.131) (105,0.131) (120,0.473) (135,0.473) (150,0.551) (165,0.473) (180,0.473) (195,1.024) (210,0.998) (225,1.050) (240,0.473) (255,0.473) (270,0.473) (285,0.473) (300,0.473)};
\addlegendentry{seed 1342}
\addplot[only marks, mark=o, mark size=2.6pt, blue!55!black, thick, nodes near coords, point meta=explicit,
  every node near coord/.style={font=\tiny, anchor=south, yshift=2pt}, forget plot]
  coordinates {(135,0.683) [0.38] (225,0.131) [0.78]};
\addplot[only marks, mark=o, mark size=2.6pt, green!45!black, thick, nodes near coords, point meta=explicit,
  every node near coord/.style={font=\tiny, anchor=south, yshift=2pt}, forget plot]
  coordinates {(105,0.473) [0.29] (240,0.683) [0.73] (285,0.945) [0.48] (300,0.131) [0.54]};
\addplot[only marks, mark=o, mark size=2.6pt, red!65!black, thick, nodes near coords, point meta=explicit,
  every node near coord/.style={font=\tiny, anchor=south, yshift=2pt}, forget plot]
  coordinates {(135,0.473) [0.70] (195,1.024) [1.00] (210,0.998) [0.46] (225,1.050) [0.68] (300,0.473) [0.57]};
\end{axis}
\end{tikzpicture}
\Description{Six reward curves against PPO iteration with dashed horizontal lines at the
random-play level and the finite-memory ceilings; five curves rise from a plateau below the
one-symbol ceiling to the full-state optimum and then fall, one never leaves the plateau.
Circled points on the second-batch curves carry the decision-token probe accuracy.}%
\caption{Reward arcs in the shortcut regime against the finite-memory ceilings (dashed; the
lowest line is random play at $0.18$). Circled points are the probed adapters of
Table~\ref{tab:llmarc}, labeled with the decision-token linear probe (chance $0.14$--$0.20$).}
\label{fig:llmarc}
\end{figure}

\textbf{The escape dynamics appear, and the escape is transient.} To test the reward-shortcut
lock of \S\ref{sec:opt} directly, we built a cell where a policy with a short memory of the
emitted symbols earns much less than a state-tracking one. A \emph{finite-memory ceiling} $V_m$
is the best reward available to a policy that conditions only on the last $m$ emitted symbols.
We compute it exactly (released script, certified against enumeration at $m{=}1$). On
\texttt{mod\_7} at $L{=}8$ with $p_{\mathrm{control}}{=}0.8$ the ladder, as a fraction of the
full-state optimum, is random $0.18$, $V_1{=}0.71$, $V_2{=}0.76$, $V_4{=}0.96$, full $1.00$. We
then ran PPO$+$GAE on Qwen2.5-3B (LoRA rank 16, $300$ iterations, terminal task reward only) for
six seeds in two batches of three, with the cell chosen from the ladder before any PPO run; the
second batch saved the adapter every $15$ iterations. All rewards in this paragraph are fractions of the full-state optimum. Every seed
spends part of its run on a plateau at $0.47$--$0.68$, below the one-symbol ceiling, and most
also sit below random play ($0.13$) for a stretch. Five of the six climb to peaks of
$0.95$--$1.05$, at the $V_4$ ceiling and above it, up to the full-state optimum (all three
first-batch seeds between iterations $150$ and $165$; the second batch at $225$ and $285$).
Checkpoint rewards are finite-sample estimates (a value above $1.00$ is sampling noise), so one
reading above $V_4$ does not by itself establish full-state optimality. The probed peak adapters
in Table~\ref{tab:llmarc} are the direct read-out. Sustained reward above a certified memory-$m$
ceiling would rule out policies restricted to that memory class. The
escape does not hold. Four of the five peaks are followed by a collapse to $0.13$, below random
play, and the fifth returns to the plateau. The sixth seed never leaves the plateau. This is the
lock-and-escape shape of \S\ref{sec:opt}, and here the escape does not last at this budget.
Figure~\ref{fig:llmarc} draws the six arcs against the ladder.

Table~\ref{tab:llmarc} probes the saved adapters at two positions: the \emph{decision token}
(the last position, which the policy conditions its choice on) and the last \emph{stream token}
(the end of the episode log, before the final prompt line), with a linear read-out and a one-hidden-layer MLP.
The frozen base model is already decodable at the decision token on
this short horizon ($0.87$ against chance $0.14$), so an absolute threshold on the probe says
little here. Decodability does not track the phases monotonically: the two peak checkpoints of
seed 1342 that lie $15$ iterations apart read $1.00$ and $0.46$. Collapse does not remove the
state. Both collapsed checkpoints keep selective probes ($0.78$ at the decision token for seed
1340, $0.90$ at the stream token for seed 1341) while reward sits below random play. The policy
fails while the representation stays, which is the planning-gap side of the instrument.

\textbf{No reward-successful perception gap at this scale.} A reward-successful perception gap
at LLM scale would be
reward clearly above random with the probe at chance and capacity saturated. It did not appear
in our runs, for a reason the synthetic suite does not have. A pretrained model starts with part
of the state already decodable at the horizons where reward is learnable. Training could in
principle erase that decodability, but it did not in our runs, so the probe-at-chance leg never
occurred. At the horizons where the frozen probe is near chance, the adaptation
bound is itself partial and no optimizer we ran raises reward. The separating power of the instrument survives
in a different form. In the shortcut regime, reward and the probe disagree in both directions:
plateau reward with above-baseline decodability, and collapsed reward with the state intact. At
this scale the probe has to be read against the frozen baseline instead of chance. Within
Qwen2.5 1.5B--14B, larger models do not raise the achieved recovery at longer horizons
(Table~\ref{tab:llmcap}). We release the instrument (natural-language environment, exact oracle,
residual-stream probe with selectivity, frozen and supervised capacity bounds, the ceiling
script, and both RL loops).

% Body generated by scripts/gen_llmcap_table.py from runs/llm/supcap2_*.json.
\begin{table}[h]\centering\small
\caption{LLM-agent supervised recovery under LoRA adaptation (Qwen2.5-Instruct, fine-tuned to
predict $q_t$; fraction of the achievable balanced-accuracy gap recovered on held-out episodes;
one run per cell).}
\label{tab:llmcap}
\begin{tabular}{llccc}
\toprule
task & model & $L{=}8$ & $L{=}16$ & $L{=}24$\\
\midrule
\texttt{parity} ($k{=}2$) & 1.5B & $0.88$ & $0.40$ & $0.32$\\
 & 3B & $0.68$ & $0.54$ & $0.23$\\
 & 7B & $0.90$ & $0.41$ & $0.34$\\
 & 14B & $0.66$ & $0.30$ & $0.64$\\
\midrule
\texttt{mod\_3} ($k{=}3$) & 1.5B & $0.51$ & $0.31$ & $0.25$\\
 & 3B & $0.48$ & $0.33$ & $0.23$\\
 & 7B & $0.60$ & $0.28$ & $0.23$\\
 & 14B & $0.56$ & $0.40$ & $0.25$\\
\midrule
\texttt{mod\_5} ($k{=}5$) & 1.5B & $0.52$ & $0.21$ & $0.19$\\
 & 3B & $0.53$ & $0.25$ & $0.17$\\
 & 7B & $0.57$ & $0.26$ & $0.18$\\
 & 14B & $0.52$ & $0.34$ & $0.19$\\
\midrule
\texttt{mod\_7} ($k{=}7$) & 1.5B & $0.56$ & $0.22$ & $0.13$\\
 & 3B & $0.49$ & $0.30$ & $0.16$\\
 & 7B & $0.56$ & $0.26$ & $0.13$\\
 & 14B & $0.62$ & $0.23$ & $0.17$\\
\bottomrule
\end{tabular}
\end{table}

% Body generated by scripts/gen_shortcut_probe_table.py from
% runs/llm/shortcut_ppo2_m7L8c8_s*/{history,adapter_probes_v2}.json.
\begin{table}[h]\centering\small
\caption{Shortcut-regime probes over saved adapters (Qwen2.5-3B, \texttt{mod\_7}, $L{=}8$,
$p_{\mathrm{control}}{=}0.8$, second batch). Phase is by the eval reward concurrent with the
save (plateau $0.40$--$0.88$, peak ${\ge}0.90$, collapsed ${\le}0.30$); reward is the fraction of
the full-state optimum under re-evaluation. Probes are balanced accuracy (chance
$0.14$--$0.20$) at the decision token (linear) and the last stream token (linear and MLP).}
\label{tab:llmarc}
\begin{tabular}{llcrccc}
\toprule
seed & iteration & phase & reward & decision & stream & stream MLP\\
\midrule
1340 & 135 & plateau & $0.66$ & $0.38$ & $0.30$ & $0.49$\\
 & 225 & collapsed & $0.06$ & $0.78$ & $0.80$ & $0.21$\\
\midrule
1341 & 105 & plateau & $0.50$ & $0.29$ & $0.37$ & $0.26$\\
 & 240 & plateau & $0.61$ & $0.73$ & $0.47$ & $0.73$\\
 & 285 & peak & $0.87$ & $0.48$ & $0.27$ & $0.46$\\
 & 300 & collapsed & $0.06$ & $0.54$ & $0.81$ & $0.90$\\
\midrule
1342 & frozen & base & $0.21$ & $0.87$ & $0.85$ & $0.85$\\
 & 135 & plateau & $0.50$ & $0.70$ & $0.38$ & $0.75$\\
 & 195 & peak & $0.96$ & $1.00$ & $0.80$ & $1.00$\\
 & 210 & peak & $0.86$ & $0.46$ & $0.45$ & $0.63$\\
 & 225 & peak & $0.95$ & $0.68$ & $0.62$ & $0.99$\\
 & 300 & plateau & $0.50$ & $0.57$ & $0.46$ & $0.24$\\
\bottomrule
\end{tabular}
\end{table}

\section{Task catalog}\label{app:catalog}
Table~\ref{tab:catalog} lists the 15 curated tasks with their structural properties, generated
directly from the released code (\texttt{scripts/gen\_task\_table.py}). Shared protocol unless a
section states otherwise: $p_{\mathrm{control}}=0.5$; training lengths $4$--$32$ with OOD test
lengths $65$--$128$; window $K{=}8$ for the window-limited encoders ($K{=}16$ for the pilot
tasks); per-stage architectures, budgets, and seeds are in the released README. The mined tasks
are RPNI reconstructions from labeled samples of our implementations of the deployed checksum
rules (\S\ref{sec:mined}).

\begin{catalogtable}
\caption{The curated task catalog. $L$ is the control leverage at $T{=}32$,
$p_{\mathrm{control}}{=}0.5$; permutation is the no-merging property of \S\ref{sec:predictor}.}
\label{tab:catalog}
\begin{tabular}{lcccc}
\toprule
task & $|Q|$ & $|\Sigma|$ & permutation & $L$\\
\midrule
\texttt{parity} & 2 & 2 & yes & 0.50\\
\texttt{mod\_2} & 2 & 2 & no & 0.50\\
\texttt{mod\_3} & 3 & 2 & yes & 0.54\\
\texttt{mod\_5} & 5 & 2 & yes & 0.43\\
\texttt{mod\_7} & 7 & 2 & yes & 0.41\\
\texttt{mod\_9} & 9 & 2 & yes & 0.35\\
\texttt{mod\_11} & 11 & 2 & yes & 0.33\\
\texttt{no\_three\_zeros} & 4 & 2 & no & 0.66\\
\texttt{ends\_with\_101} & 4 & 2 & no & 0.34\\
\texttt{ends\_with\_1000} & 5 & 2 & no & 0.27\\
\texttt{workflow\_api} & 6 & 4 & no & 0.71\\
\texttt{workflow\_longrange} & 5 & 5 & no & 0.50\\
\texttt{workflow\_reconcile} & 5 & 3 & yes & 0.52\\
\texttt{mined\_div7} & 7 & 2 & yes & 0.41\\
\texttt{mined\_ean} & 20 & 10 & yes & 0.50\\
\bottomrule
\end{tabular}
\end{catalogtable}

\section{AI use disclosure}\label{app:aiuse}
Generative AI tools were used in this work, in every case under the authors' direction and review.
AI helped create and edit the code and scripts for the experiments (building on the authors'
transformer training library, which was written by the authors) and helped draft, edit, and format
the text of the paper. Every figure is rendered from the tracked run outputs by project plotting
code; no generative image model produced any figure or other media.

AI contributed to hypotheses and methodology in three ways, through GPT-5-series assistant
models; per-session records were not kept, so for the earlier sessions the exact version
(GPT-5.5 or GPT-5.6 Sol) is not recorded. First, a GPT-5-series model helped design and
give feedback on the methodology and experiments and helped propose and refine hypotheses,
including the pre-specified pair of Appendix~\ref{app:prereg}. This was interactive
multi-turn assistance over the course of the project. No single prompt was stored, and none is
reconstructed here. The standing request in those sessions was to critique the experimental
design and propose controls and alternative explanations. Second, a GPT-5-series model
helped formulate the mathematical claims of Appendix~\ref{app:credit} and assisted with their
proofs. A near-final draft of that appendix was reviewed by GPT-5.6 Sol under the standing
request to check the manuscript for errors. The authors re-derived the surviving proofs by hand.
Third, AI found and formatted references, and the authors verified each against its source.

No AI-assisted cleaning or reformatting took place: the synthetic suites are generated by project
code, and the deterministic preprocessing of the public event logs is documented in the released
code. The authors reviewed all AI-assisted work,
checked every number in the paper against the tracked run outputs, re-ran the analysis pipeline end
to end from the released code, and accept full responsibility for the accuracy, originality, and
citations of this paper.


\small
\begin{thebibliography}{99}
\bibitem[Alain \& Bengio(2017)]{alain2016} Guillaume Alain and Yoshua Bengio. 2017. Understanding Intermediate Layers Using Linear Classifier Probes. In \emph{5th International Conference on Learning Representations (ICLR 2017), Workshop Track}. \url{https://arxiv.org/abs/1610.01644}
\bibitem[Angluin(1987)]{angluin1987} Dana Angluin. 1987. Learning Regular Sets from Queries and Counterexamples. \emph{Information and Computation} 75, 2 (1987), 87--106. \url{https://doi.org/10.1016/0890-5401(87)90052-6}
\bibitem[Bhattamishra et al.(2020)]{bhattamishra2020} Satwik Bhattamishra, Kabir Ahuja, and Navin Goyal. 2020. On the Ability and Limitations of Transformers to Recognize Formal Languages. In \emph{Proceedings of the 2020 Conference on Empirical Methods in Natural Language Processing (EMNLP 2020)}. Association for Computational Linguistics, 7096--7116. \url{https://doi.org/10.18653/v1/2020.emnlp-main.576}
\bibitem[Brafman \& De Giacomo(2019)]{brafman2019rdp} Ronen I. Brafman and Giuseppe De Giacomo. 2019. Regular Decision Processes: A Model for Non-Markovian Domains. In \emph{Proceedings of the Twenty-Eighth International Joint Conference on Artificial Intelligence (IJCAI 2019)}. 5516--5522. \url{https://doi.org/10.24963/ijcai.2019/766}
\bibitem[Chang \& Bisk(2025)]{chang2024count} Yingshan Chang and Yonatan Bisk. 2025. Language Models Need Inductive Biases to Count Inductively. In \emph{The Thirteenth International Conference on Learning Representations (ICLR 2025)}. \url{https://arxiv.org/abs/2405.20131}
\bibitem[Cho et al.(2014)]{cho2014gru} Kyunghyun Cho, Bart van Merri{\"e}nboer, Caglar Gulcehre, Dzmitry Bahdanau, Fethi Bougares, Holger Schwenk, and Yoshua Bengio. 2014. Learning Phrase Representations using {RNN} Encoder--Decoder for Statistical Machine Translation. In \emph{Proceedings of the 2014 Conference on Empirical Methods in Natural Language Processing (EMNLP 2014)}. Association for Computational Linguistics, 1724--1734. \url{https://doi.org/10.3115/v1/D14-1179}
\bibitem[de Haan et al.(2019)]{dehaan2019} Pim de Haan, Dinesh Jayaraman, and Sergey Levine. 2019. Causal Confusion in Imitation Learning. In \emph{Advances in Neural Information Processing Systems 32 (NeurIPS 2019)}. Curran Associates, Inc., 11693--11704. \url{https://arxiv.org/abs/1905.11979}
\bibitem[de Leoni \& Mannhardt(2015)]{roadfines} Massimiliano de Leoni and Felix Mannhardt. 2015. Road Traffic Fine Management Process. Dataset, 4TU.ResearchData. \url{https://doi.org/10.4121/uuid:270fd440-1057-4fb9-89a9-b699b47990f5}
\bibitem[Del{\'e}tang et al.(2023)]{deletang2023} Gr{\'e}goire Del{\'e}tang, Anian Ruoss, Jordi Grau-Moya, Tim Genewein, Li Kevin Wenliang, Elliot Catt, Chris Cundy, Marcus Hutter, Shane Legg, Joel Veness, and Pedro A. Ortega. 2023. Neural Networks and the {Chomsky} Hierarchy. In \emph{The Eleventh International Conference on Learning Representations (ICLR 2023)}. \url{https://arxiv.org/abs/2207.02098}
\bibitem[Geirhos et al.(2020)]{geirhos2020shortcut} Robert Geirhos, J{\"o}rn-Henrik Jacobsen, Claudio Michaelis, Richard Zemel, Wieland Brendel, Matthias Bethge, and Felix A. Wichmann. 2020. Shortcut Learning in Deep Neural Networks. \emph{Nature Machine Intelligence} 2, 11 (2020), 665--673. \url{https://doi.org/10.1038/s42256-020-00257-z}
\bibitem[Gelada et al.(2019)]{gelada2019} Carles Gelada, Saurabh Kumar, Jacob Buckman, Ofir Nachum, and Marc G. Bellemare. 2019. {DeepMDP}: Learning Continuous Latent Space Models for Representation Learning. In \emph{Proceedings of the 36th International Conference on Machine Learning (ICML 2019)} (Proceedings of Machine Learning Research, Vol. 97). PMLR, 2170--2179. \url{https://arxiv.org/abs/1906.02736}
\bibitem[Hahn(2020)]{hahn2020limits} Michael Hahn. 2020. Theoretical Limitations of Self-Attention in Neural Sequence Models. \emph{Transactions of the Association for Computational Linguistics} 8 (2020), 156--171. \url{https://doi.org/10.1162/tacl_a_00306}
\bibitem[Hewitt \& Liang(2019)]{hewitt2019} John Hewitt and Percy Liang. 2019. Designing and Interpreting Probes with Control Tasks. In \emph{Proceedings of the 2019 Conference on Empirical Methods in Natural Language Processing and the 9th International Joint Conference on Natural Language Processing (EMNLP-IJCNLP 2019)}. Association for Computational Linguistics, 2733--2743. \url{https://doi.org/10.18653/v1/D19-1275}
\bibitem[Hu et al.(2022)]{hu2022lora} Edward J. Hu, Yelong Shen, Phillip Wallis, Zeyuan Allen-Zhu, Yuanzhi Li, Shean Wang, Lu Wang, and Weizhu Chen. 2022. {LoRA}: Low-Rank Adaptation of Large Language Models. In \emph{The Tenth International Conference on Learning Representations (ICLR 2022)}. \url{https://arxiv.org/abs/2106.09685}
\bibitem[X. Huang et al.(2025)]{huang2025lengthgen} Xinting Huang, Andy Yang, Satwik Bhattamishra, Yash Sarrof, Andreas Krebs, Hattie Zhou, Preetum Nakkiran, and Michael Hahn. 2025. A Formal Framework for Understanding Length Generalization in Transformers. In \emph{The Thirteenth International Conference on Learning Representations (ICLR 2025)}. \url{https://arxiv.org/abs/2410.02140}
\bibitem[Y. Huang et al.(2025)]{cot2025} Yu Huang, Zixin Wen, Aarti Singh, Yuejie Chi, and Yuxin Chen. 2025. Transformers Provably Learn Chain-of-Thought Reasoning with Length Generalization. In \emph{Advances in Neural Information Processing Systems 38 (NeurIPS 2025)}. Curran Associates, Inc. \url{https://doi.org/10.52202/085713-4230}
\bibitem[Jaderberg et al.(2017)]{jaderberg2017unreal} Max Jaderberg, Volodymyr Mnih, Wojciech Marian Czarnecki, Tom Schaul, Joel Z. Leibo, David Silver, and Koray Kavukcuoglu. 2017. Reinforcement Learning with Unsupervised Auxiliary Tasks. In \emph{The Fifth International Conference on Learning Representations (ICLR 2017)}. \url{https://arxiv.org/abs/1611.05397}
\bibitem[Jin \& Rinard(2024)]{jin2024programs} Charles Jin and Martin Rinard. 2024. Emergent Representations of Program Semantics in Language Models Trained on Programs. In \emph{Proceedings of the 41st International Conference on Machine Learning (ICML 2024)} (Proceedings of Machine Learning Research, Vol. 235). PMLR, 22160--22184. \url{https://arxiv.org/abs/2305.11169}
\bibitem[Karvonen(2024)]{karvonen2024chess} Adam Karvonen. 2024. Emergent World Models and Latent Variable Estimation in Chess-Playing Language Models. In \emph{First Conference on Language Modeling (COLM 2024)}. \url{https://arxiv.org/abs/2403.15498}
\bibitem[Krohn \& Rhodes(1965)]{krohn1965} Kenneth Krohn and John Rhodes. 1965. Algebraic Theory of Machines. {I}. Prime Decomposition Theorem for Finite Semigroups and Machines. \emph{Transactions of the American Mathematical Society} 116 (1965), 450--464. \url{https://doi.org/10.1090/S0002-9947-1965-0188316-1}
\bibitem[Langosco Di Langosco et al.(2022)]{langosco2022} Lauro Langosco Di Langosco, Jack Koch, Lee D. Sharkey, Jacob Pfau, and David Krueger. 2022. Goal Misgeneralization in Deep Reinforcement Learning. In \emph{Proceedings of the 39th International Conference on Machine Learning (ICML 2022)} (Proceedings of Machine Learning Research, Vol. 162). PMLR, 12004--12019. \url{https://arxiv.org/abs/2105.14111}
\bibitem[Li et al.(2023)]{li2023othello} Kenneth Li, Aspen K. Hopkins, David Bau, Fernanda Vi{\'e}gas, Hanspeter Pfister, and Martin Wattenberg. 2023. Emergent World Representations: Exploring a Sequence Model Trained on a Synthetic Task. In \emph{The Eleventh International Conference on Learning Representations (ICLR 2023)}. \url{https://arxiv.org/abs/2210.13382}
\bibitem[Littman et al.(2001)]{littman2001psr} Michael L. Littman, Richard S. Sutton, and Satinder Singh. 2001. Predictive Representations of State. In \emph{Advances in Neural Information Processing Systems 14 (NIPS 2001)}. MIT Press, 1555--1561.
\bibitem[Liu et al.(2023)]{liu2022shortcuts} Bingbin Liu, Jordan T. Ash, Surbhi Goel, Akshay Krishnamurthy, and Cyril Zhang. 2023. Transformers Learn Shortcuts to Automata. In \emph{The Eleventh International Conference on Learning Representations (ICLR 2023)}. \url{https://arxiv.org/abs/2210.10749}
\bibitem[Mannhardt(2016)]{sepsis} Felix Mannhardt. 2016. Sepsis Cases -- Event Log. Dataset, 4TU.ResearchData. \url{https://doi.org/10.4121/uuid:915d2bfb-7e84-49ad-a286-dc35f063a460}
\bibitem[Mannhardt(2017)]{hospitalbilling} Felix Mannhardt. 2017. Hospital Billing -- Event Log. Dataset, 4TU.ResearchData. \url{https://doi.org/10.4121/uuid:76c46b83-c930-4798-a1c9-4be94dfeb741}
\bibitem[Michalenko et al.(2019)]{michalenko2019} Joshua J. Michalenko, Ameesh Shah, Abhinav Verma, Richard G. Baraniuk, Swarat Chaudhuri, and Ankit B. Patel. 2019. Representing Formal Languages: A Comparison Between Finite Automata and Recurrent Neural Networks. In \emph{The Seventh International Conference on Learning Representations (ICLR 2019)}. \url{https://arxiv.org/abs/1902.10297}
\bibitem[Mnih et al.(2016)]{mnih2016a3c} Volodymyr Mnih, Adria Puigdomenech Badia, Mehdi Mirza, Alex Graves, Timothy Lillicrap, Tim Harley, David Silver, and Koray Kavukcuoglu. 2016. Asynchronous Methods for Deep Reinforcement Learning. In \emph{Proceedings of the 33rd International Conference on Machine Learning (ICML 2016)} (Proceedings of Machine Learning Research, Vol. 48). PMLR, 1928--1937. \url{https://arxiv.org/abs/1602.01783}
\bibitem[Nanda et al.(2023)]{nanda2023linear} Neel Nanda, Andrew Lee, and Martin Wattenberg. 2023. Emergent Linear Representations in World Models of Self-Supervised Sequence Models. In \emph{Proceedings of the 6th BlackboxNLP Workshop: Analyzing and Interpreting Neural Networks for NLP (BlackboxNLP 2023)}. Association for Computational Linguistics, 16--30. \url{https://doi.org/10.18653/v1/2023.blackboxnlp-1.2}
\bibitem[Ng et al.(1999)]{ng1999shaping} Andrew Y. Ng, Daishi Harada, and Stuart Russell. 1999. Policy Invariance Under Reward Transformations: Theory and Application to Reward Shaping. In \emph{Proceedings of the Sixteenth International Conference on Machine Learning (ICML 1999)}. Morgan Kaufmann, 278--287.
\bibitem[Ni et al.(2022)]{ni2022} Tianwei Ni, Benjamin Eysenbach, and Ruslan Salakhutdinov. 2022. Recurrent Model-Free {RL} Can Be a Strong Baseline for Many {POMDPs}. In \emph{Proceedings of the 39th International Conference on Machine Learning (ICML 2022)} (Proceedings of Machine Learning Research, Vol. 162). PMLR, 16691--16723. \url{https://arxiv.org/abs/2110.05038}
\bibitem[Oncina \& Garc{\'\i}a(1992)]{oncina1992} Jos{\'e} Oncina and Pedro Garc{\'\i}a. 1992. Inferring Regular Languages in Polynomial Updated Time. In \emph{Pattern Recognition and Image Analysis} (Nicol{\'a}s P{\'e}rez de la Blanca, Alberto Sanfeliu, and Enrique Vidal (Eds.)). World Scientific, 49--61. \url{https://doi.org/10.1142/9789812797902_0004}
\bibitem[Pan et al.(2022)]{pan2022mis} Alexander Pan, Kush Bhatia, and Jacob Steinhardt. 2022. The Effects of Reward Misspecification: Mapping and Mitigating Misaligned Models. In \emph{The Tenth International Conference on Learning Representations (ICLR 2022)}. \url{https://arxiv.org/abs/2201.03544}
\bibitem[Qwen Team(2024)]{qwen2025} Qwen Team. 2024. Qwen2.5 Technical Report. arXiv preprint. \url{https://arxiv.org/abs/2412.15115}
\bibitem[Schulman et al.(2016)]{schulman2016gae} John Schulman, Philipp Moritz, Sergey Levine, Michael Jordan, and Pieter Abbeel. 2016. High-Dimensional Continuous Control Using Generalized Advantage Estimation. In \emph{The Fourth International Conference on Learning Representations (ICLR 2016)}. \url{https://arxiv.org/abs/1506.02438}
\bibitem[Schulman et al.(2017)]{schulman2017ppo} John Schulman, Filip Wolski, Prafulla Dhariwal, Alec Radford, and Oleg Klimov. 2017. Proximal Policy Optimization Algorithms. arXiv preprint. \url{https://arxiv.org/abs/1707.06347}
\bibitem[Sch{\"u}tzenberger(1965)]{schutzenberger1965} M. P. Sch{\"u}tzenberger. 1965. On Finite Monoids Having Only Trivial Subgroups. \emph{Information and Control} 8, 2 (1965), 190--194. \url{https://doi.org/10.1016/S0019-9958(65)90108-7}
\bibitem[Shah et al.(2022)]{shah2022} Rohin Shah, Vikrant Varma, Ramana Kumar, Mary Phuong, Victoria Krakovna, Jonathan Uesato, and Zac Kenton. 2022. Goal Misgeneralization: Why Correct Specifications Aren't Enough For Correct Goals. arXiv preprint. \url{https://arxiv.org/abs/2210.01790}
\bibitem[Skalse et al.(2022)]{skalse2022} Joar Skalse, Nikolaus Howe, Dmitrii Krasheninnikov, and David Krueger. 2022. Defining and Characterizing Reward Gaming. In \emph{Advances in Neural Information Processing Systems 35 (NeurIPS 2022)}. Curran Associates, Inc., 9460--9471. \url{https://doi.org/10.52202/068431-0687}
\bibitem[Strobl et al.(2024)]{strobl2024survey} Lena Strobl, William Merrill, Gail Weiss, David Chiang, and Dana Angluin. 2024. What Formal Languages Can Transformers Express? A Survey. \emph{Transactions of the Association for Computational Linguistics} 12 (2024), 543--561. \url{https://doi.org/10.1162/tacl_a_00663}
\bibitem[Toro Icarte et al.(2018)]{icarte2018} Rodrigo Toro Icarte, Toryn Klassen, Richard Valenzano, and Sheila McIlraith. 2018. Using Reward Machines for High-Level Task Specification and Decomposition in Reinforcement Learning. In \emph{Proceedings of the 35th International Conference on Machine Learning (ICML 2018)} (Proceedings of Machine Learning Research, Vol. 80). PMLR, 2107--2116.
\bibitem[Toro Icarte et al.(2023)]{icarte2022} Rodrigo Toro Icarte, Toryn Q. Klassen, Richard Valenzano, Margarita P. Castro, Ethan Waldie, and Sheila A. McIlraith. 2023. Learning Reward Machines: A Study in Partially Observable Reinforcement Learning. \emph{Artificial Intelligence} 323 (2023), 103989. \url{https://doi.org/10.1016/j.artint.2023.103989}
\bibitem[Toshniwal et al.(2022)]{toshniwal2022chess} Shubham Toshniwal, Sam Wiseman, Karen Livescu, and Kevin Gimpel. 2022. Chess as a Testbed for Language Model State Tracking. In \emph{Proceedings of the AAAI Conference on Artificial Intelligence (AAAI 2022)} (Vol. 36, No. 10). 11385--11393. \url{https://doi.org/10.1609/aaai.v36i10.21390}
\bibitem[van Dongen(2012)]{bpic12} Boudewijn van Dongen. 2012. BPI Challenge 2012. Dataset, 4TU.ResearchData. \url{https://doi.org/10.4121/uuid:3926db30-f712-4394-aebc-75976070e91f}
\bibitem[van Dongen(2015)]{bpic15} Boudewijn van Dongen. 2015. BPI Challenge 2015 Municipality 1. Dataset, 4TU.ResearchData. \url{https://doi.org/10.4121/uuid:a0addfda-2044-4541-a450-fdcc9fe16d17}
\bibitem[van Dongen(2017)]{bpic17} Boudewijn van Dongen. 2017. BPI Challenge 2017. Dataset, 4TU.ResearchData. \url{https://doi.org/10.4121/uuid:5f3067df-f10b-45da-b98b-86ae4c7a310b}
\bibitem[van Dongen(2019)]{bpic19} Boudewijn van Dongen. 2019. BPI Challenge 2019. Dataset, 4TU.ResearchData. \url{https://doi.org/10.4121/uuid:d06aff4b-79f0-45e6-8ec8-e19730c248f1}
\bibitem[Weiss et al.(2018)]{weiss2018} Gail Weiss, Yoav Goldberg, and Eran Yahav. 2018. On the Practical Computational Power of Finite Precision {RNNs} for Language Recognition. In \emph{Proceedings of the 56th Annual Meeting of the Association for Computational Linguistics (ACL 2018), Volume 2: Short Papers}. Association for Computational Linguistics, 740--745. \url{https://doi.org/10.18653/v1/P18-2117}
\bibitem[Zhang et al.(2021)]{zhang2021dbc} Amy Zhang, Rowan McAllister, Roberto Calandra, Yarin Gal, and Sergey Levine. 2021. Learning Invariant Representations for Reinforcement Learning without Reconstruction. In \emph{The Ninth International Conference on Learning Representations (ICLR 2021)}. \url{https://arxiv.org/abs/2006.10742}
\end{thebibliography}
\end{document}